\documentclass[final]{style}

\usepackage{times}
\usepackage{epsfig}
\usepackage{graphicx}
\usepackage{amsmath}
\usepackage{amssymb}
\usepackage{caption}
\usepackage{placeins}
\usepackage{multirow}
\usepackage[normalem]{ulem}

\usepackage{hyperref}
\hypersetup{colorlinks,bookmarks=false,breaklinks=true,pagebackref=true}

\begin{document}
\title{Towards General Purpose, Geometry Preserving Single-View Depth Estimation}
\author{Mikhail Romanov\\
{\tt\small m.romanov@samsung.com}
\and
Nikolay Patakin\\
{\tt\small n.patakin@samsung.com}
\and
Anna Vorontsova\\
{\tt\small a.vorontsova@samsung.com}
\and
Sergey Nikolenko\\
{\tt\small s.nikolenko@samsung.com}
\and
Dmitriy Senyushkin\\
{\tt\small d.senyushkin@samsung.com}
\and
Anton Konushin\\
{\tt\small a.konushin@samsung.com}
\\
}

\twocolumn[{
    \renewcommand\twocolumn[1][]{#1}
    \maketitle
    \centering
    \includegraphics[width=\textwidth]{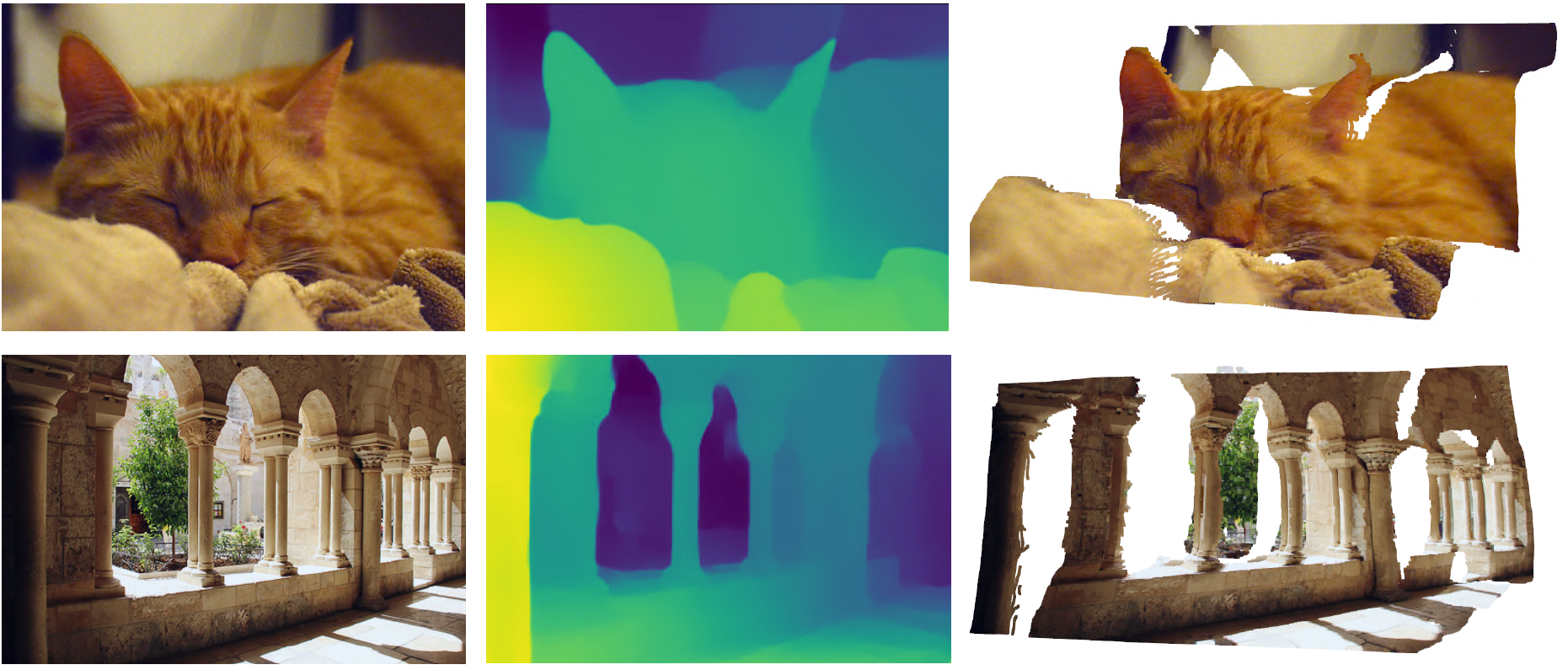}
    \captionof{figure}{Left to right: original image (from the \emph{Depth-in-the-Wild dataset}, unseen during training), corresponding up-to-scale inverse depth map predicted with the proposed B5-LRN model, and a point cloud built directly from this depth map.}
    \label{fig:cover}
    \vspace{.3cm}
}]

\begin{abstract}

Single-view depth estimation (SVDE) plays a crucial role in scene understanding for AR applications, 3D modeling, and robotics, providing the geometry of a scene based on a single image. Recent works have shown that a successful solution strongly relies on the diversity and volume of training data. This data can be sourced from stereo movies and photos. However, they do not provide geometrically complete depth maps (as disparities contain unknown shift value). Therefore, existing models trained on this data are not able to recover correct 3D representations. Our work shows that a model trained on this data along with conventional datasets can gain accuracy while predicting correct scene geometry. Surprisingly, only a small portion of geometrically correct depth maps are required to train a model that performs equally to a model trained on the full geometrically correct dataset. After that, we train computationally efficient models on a mixture of datasets using the proposed method. Through quantitative comparison on completely unseen datasets and qualitative comparison of 3D point clouds, we show that our model defines the new state of the art in general-purpose SVDE.

\end{abstract}

\section{Introduction}

Single-view monocular depth estimation (SVDE) is an essential problem of computer vision and visual understanding. It has numerous important applications in such areas as simultaneous localization and mapping (SLAM), navigation, computational photography, robotics, augmented reality, and many others. Approaches capable of predicting geometrically correct depth maps are especially interesting for industrial applications as they would potentially enable the users to construct a reliable three-dimensional point cloud based on a single image. Unfortunately, even state-of-the-art approaches cannot either provide correct scene geometry or perform robustly on arbitrary images.

Nevertheless, recent developments based on deep neural networks have led to significant progress in this area over the last few years; see, e.g., the survey~\cite{Zhao_2020}. Up until very recently, deep learning models had usually been trained and tested either on synthetic datasets such as SYNTHIA~\cite{synthia} or SunCG~\cite{suncg} or datasets with sensor-based or geometry-based depth supervision, such as KITTI~\cite{kitti2015}, NYU~\cite{nyuv2}, and others (we discuss existing depth datasets in detail in Section~\ref{sec:depth}).

From the other side, several recent works improved depth estimation accuracy through finding new data sources. Among them, stereo movies \cite{midas, wang2019web} and photos \cite{redweb} are one of the most diverse and accessible at-scale data sources. The depth data available in stereo movies is voluminous and diverse, but in order to derive from them geometrically complete depth data one would have to know precise intrinsic and extrinsic parameters of the stereo cameras. These parameters are not needed to consume stereo movies and thus are usually not provided. Without them, disparity from a stereo pair can be computed up to unknown shift and scale coefficients (UTSS). Such information can be used as a good proxy for depth, but is insufficient to restore geometry. Due to this fact, state-of-the-art models that are currently trained using data from stereo movies can show good performance and generalization properties but do not provide geometrically correct predictions \cite{midas, wang2019web, redweb}, 

In this work, we show that it is possible to use stereo movies data even without precise camera parameters in training networks that predict geometrically correct depth maps. Moreover, we show that use of this large-scale source of depth data from stereo movies alongside with a small quantity of geometrically correct depth data (provided by sensors or structure-from-motion reconstructions) in the training process of an SVDE model is equivalent to training the same model using a dataset of the same size but containing only geometrically correct data. Inspiring by Eigen et al. \cite{eigen}, we introduce a new scale-invariant pairwise loss function, which outperforms existing data terms on NYU \cite{nyuv2}. 

Since depth estimation problem has many industrial applications, which means that a real-life depth estimation network should be not only accurate but also computationally efficient, we construct our solutions using MobileNet~\cite{mobilenet,mobilenetv2} and EfficientNet~\cite{efficientnet} networks as backbones and a modified Light-Weight Refine Net~\cite{nekrasov2018real} as a decoder.

Following~\cite{midas}, we train our models on a mixture of datasets, including DIML \cite{diml}, MegaDepth \cite{megadepth}, RedWeb \cite{redweb} and stereo movies, but preserve geometrically correct predictions. We test the resulting models on datasets that have not been used for training (NYU\cite{nyuv2}, TUM~\cite{tumrgbd}, ETH3D~\cite{eth3d}, DIW~\cite{diw}). Our most accurate model (B5-LRN) ourperforms MIDAS \cite{midas}, while having 3.6x less parameters. Our fastest model (based on MobilenetV2 \cite{mobilenetv2}) with only 2.4 million parameters can produce plausible 3D geometry on a wide range of scenarios (see Fig. \ref{fig:meshes}). 

\paragraph{Our contribution.}
Firstly, we propose how models can be trained on both geometrically complete and geometrically incomplete data sources without loss of ability to predict correct scene geometry. Secondly, we propose a new loss function, which outperforms existing on a NYU \cite{nyuv2} dataset. We train a set of scalable by computational complexity models. The most accurate one (B5-LRN) outperforms competitors, being a new state-of-the-art in the general-purpose SVDE. Also, we show that the small network based on MobilenetV2\cite{mobilenetv2} can still generalize well in a wide variety of scenarios. 
 
The paper is organized as follows. Section~\ref{sec:related} discusses related work. In Section~\ref{sec:depth}, we discuss existing datasets for depth estimation and three different kinds of available depth maps. In Section~\ref{sec:methods} we propose geometry-preserving training method and a new loss function, Section~\ref{sec:eval} presents the experimental results, and Section~\ref{sec:concl} concludes the paper.

\section{Related Work}\label{sec:related}

\paragraph{Single-View Depth Estimation} (SVDE) was studied for decades. Early methods of general purpose depth estimation from single RGB image applied complicated heuristic algorithms with hand-crafted features \cite{saxena,hoiem2005}.

Recently, deep learning-based approaches were adopted for solving various computer vision tasks including depth estimation. The majority of modern approaches formulate depth estimation as a dense labelling in continuous space \cite{eigen2015predicting,laina2016deeper,cao2017estimating,nekrasov2018real}. In that case, $L_1$- or $L_2$-based regression loss functions are used in different domains (depth, log-depth, disparity). In Eigen et al. \cite{eigen} it is proposed to compare pairwise differences of ground-truth and prediction. Since it computed in log-depth domain, loss is invariant to scaling. However, alternative formulations have also been considered: for instance, Fu et al. \cite{dorn} proposed to discretize depth and to interpret depth estimation as ordinal regression problem. 

While the aforementioned works mainly deal with training models using sensor-measured depth, some other works employ hand-labeled data \cite{chen2016single}, processed videos \cite{midas, wang2019web} and stereo movies \cite{midas} data for training. While being trained on such data, models learn only ordinal depth rankings \cite{diw, redweb} or predicts disparity up to unknown scale and shift coefficients (UTSS) \cite{wang2019web, midas}.

\begin{figure*}[!h]
    \centering
    \includegraphics[width=0.98\textwidth]{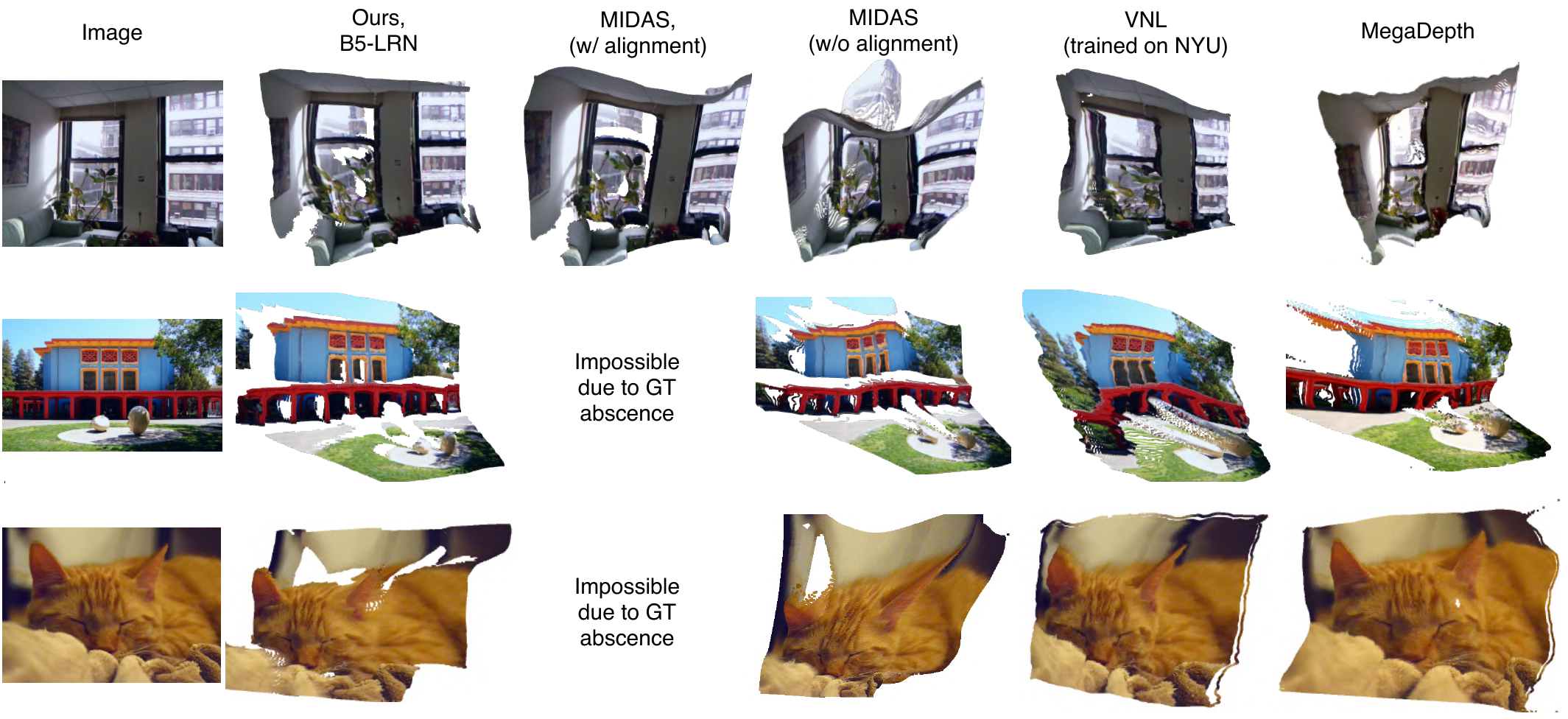}
    \caption{Point clouds constructed with depth estimation models. Top row: an image from NYU. Second and third rows: images from the DIW dataset \cite{diw}; the ground truth is not available in this case so it is impossible to align shift of MiDaS predictions.}
    \label{fig:pointclouds}\vspace{.6cm}
\end{figure*}

\section{Depth Data and Depth Predictions}\label{sec:depth}

\paragraph*{Datasets with absolute depth measurements.}
There are many datasets and data sources for the SVDE problem, and they present different kinds of data with different restrictions. First, several datasets provide images accompanied by \emph{absolute} depth measurements, usually taken with special depth sensors such as LiDARs, time-of-flight cameras, structured light sensors, and others. Datasets that provide such depth data include KITTI~\cite{kitti2015}, NYUv2~\cite{nyuv2}, DIML~\cite{diml}, ETH3D~\cite{eth3d}, Standord 2D3DS~\cite{stanford3d}, ScanNet~\cite{scannet}, Sun3D~\cite{sun3d}, SunRGBD~\cite{sunrgbd} and others. Such datasets usually either do not provide diverse data (captures only indoor environments due to the limited sensing range) or are expensive to collect at-scale and sparse (laser or LiDAR-based scanners). 

Another possible source of depth data are \emph{synthetic} datasets. If the 3D model of a synthetic scene is available, precise depth measurements can be generated together with rendered images at virtually no additional cost. Therefore, almost all modern synthetic datasets for computer vision include depth maps for their images; examples of such datasets include Sintel~\cite{sintel} and SYNTHIA~\cite{synthia}, Falling Things~\cite{8575443} for basic objects, 
SunCG~\cite{suncg} and Habitat~\cite{habitat19arxiv} for indoor scenes,
DeepDrive~\cite{craig_quiter_2018_1248998} and ProcSy~\cite{Khan_2019_CVPR_Workshops} for outdoor scenes, PHAV~\cite{8099761} and SURREAL~\cite{DBLP:journals/corr/Varol0MMBLS17} for synthetic humans, and many more; we refer to~\cite{DBLP:journals/corr/abs-1909-11512} for a detailed overview.

\paragraph*{Up-to-scale data.}
Up-to-scale (UTS) data is a different type of data for the SVDE problem, where the depth measurements are known up to an unknown constant (scale). This means that a UTS depth map for an image $\mathbf{d}$ and an absolute depth map $\mathbf{d^*}$ are related as $\mathbf{d^*}^{-1} = C_1 \mathbf{d}^{-1}$, where $C_1$ is an unknown constant. Note that UTS depth map represents the overall geometry of the scene, which means that it is sufficient to know the UTS depth map of a scene to construct a reliable point cloud.

The most popular dataset that provides such depth data is MegaDepth~\cite{megadepth}. This dataset was collected using the Structure-from-Motion (SfM) method~\cite{sfm} from crowdsourced images of architectural sights. 

\paragraph*{Up-to-shift-and-scale data} 
(UTSS) can be derived from stereo movies and stereo photos using state of the art optical flow algorithms; this has been done in datasets such as MiDaS~\cite{midas}, RedWeb\cite{redweb} and WSVD~\cite{wang2019web}.

For an aligned stereo pair, the optical flow is usually referred to as \emph{disparity}, and the disparity map is related to the absolute depth map as follows:
$$\mathbf{d^*}^{-1} = C_1 (\mathbf{D} + C_2),$$
where $\mathbf{D}$ is the disparity map and $C_1$, $C_2$ are unknown coefficients. $C_2$ is only known if we know the displacement of the principal point between the left and right frames of a stereo pair. Note that it is impossible to get absolute or UTS depth maps without knowing $C_2$. Hence, it is impossible to build a geometrically correct point cloud from this data without a correct $C_2$. More details on this can be found in supplementary.

\paragraph{From UTSS to UTS.}
As we have seen above, if one needs geometrically correct predictions one has to predict either absolute or UTS depth maps. On the other hand, the problem of predicting absolute depth is ill-posed: it is impossible to detect the scale of a scene using only a single frame, since infinite number of scenes with different scale can be projected into the same image. To correctly estimate scene scale models are likely to learn size for different types of objects. Unfortunately, absolute depth datasets for a wide range of outdoor datasets are not available. 

Since our goal is to learn geometrically correct depth estimations, we train our models in UTS mode. UTSS data lacks only one value ($C_2$) per image to be transformed to UTS data. Using an incorrect value of $C_2$ leads to incorrect reconstruction of straight lines and angles between planes. Thus, we may try to use UTSS data along with UTS data during training, with the expectation that the neural network will learn correct 3D geometry patterns from UTS data and reuse them.

\section{Loss functions}\label{sec:methods}

\paragraph{Scale Invariant Loss.} 
There are several loss functions commonly used for the depth estimation problem. One of the most popular ones is the $L_2$ pairwise loss function~\cite{eigen2015predicting}. Interesting feature of this loss is that it takes into account all the pairs of pixels on the image. In the same time pointwise $L_1$ loss functions are more robust as they pay less attention to outliers and known to work better in SVDE then the pointwise $L_2$ loss. We introduce the proposed pairwise $L_1$ loss function below and then discuss its properties.

The pairwise $L_1$ loss function for a single image can be formulated as follows:
\begin{equation}
    \mathcal{L}_{SI} = \frac{1}{N^2} \sum_{i, j=1}^N \left| (\log d_i - \log d_j) - (\log d^*_i - \log d^*_j)\right|,
    \label{eq:l1_pairwise}
\end{equation}
where $d$ is the predicted depth, $d^*$ is the ground truth depth, $i$ and $j$ are pixel indices, and $N$ is the number of pixels in the image. 
Note that since we are subtracting logarithms of depth values, the proposed pairwise $L_1$ loss is scale-invariant (SI) so it can be used for training on both absolute and UTS inverse depth maps.

A direct calculation of this loss function would require a summation of $N^2$ terms. However, it can be computed more efficiently, in time $O(N \log N)$. To do so we first introduce the differences between logarithmic depths $R_i = \log d_i - \log d_i^*$. In the terms of the differences $R_i$, loss function~\eqref{eq:l1_pairwise} can be rewritten as
\begin{equation}
    \mathcal{L}_{SI} = -\frac{2}{N^2} \sum_{i = 1}^N R_{\{i\}} \left(N - 1 - 2(i - 1) \right),
    \label{eq:l1_pairwise_fast}
\end{equation}
where $R_{\{i\}}$ is a sorted list of $R_i$: $R_{\{i\}} \geq R_{\{j\}}$ if $i > j$.
To sort the list of differences, we need $O(N \log N)$ operations, and then $\mathcal{L}_{SI}$ is computed in linear time, getting the overall computational cost for calculating the pairwise $L_1$ loss of $O(N \log N)$. A full derivation of this loss function is shown in supplementary. Despite the increased asymptotic complexity of the loss function computation, in practice we observe no more than 5\% increase in training time.

\paragraph{Shift-and-Scale Invariant Loss.}

A scale-invariant (SI) loss can be easily converted into a shift-and-scale invariant (SSI) loss. We can replace the logarithm of the depth with normalized disparity:
$$\tilde{D}_i = \frac{D_i - \mu}{\sigma},$$
where $D_i$ is the disparity value at pixel $i$, and $\mu$ and $\sigma$ are the mean and standard deviation for the image's disparity map:
$$
    \mu = \frac{1}{N} \sum_{i=1}^N D_i; \quad
    \sigma = \frac{1}{N-1} \sum_{i=1}^N (D_i - \mu)^2.
$$
The ground truth disparity map is defined by the optical flow between left and right frames, while the predicted disparity map can be computed from the depth map as $D_i = 1/d_i$.

With this in mind, we can define the SSI pairwise $L_1$ loss:
$$
    \mathcal{L}_{SSI} = \frac{1}{N^2} \sum_{i, j=1}^N | (\tilde{D}_i - \tilde{D}_j) - (\tilde{D}^*_i - \tilde{D}^*_j)|.$$
This loss function also can be computed in time $O(N \log N)$ with formula~\eqref{eq:l1_pairwise_fast}, substituting $D_i - D^*_i$ instead of $R_i$.

\paragraph{Mixing datasets.} To train on UTS, absolute, and UTSS data at the same time, we propose to use a mixture of SSI and SI loss functions, using the latter when it is available. Formally, we train our models with the following loss function:
\begin{equation}
    \mathcal{L}_{Mixture} = \mathbb{I}_{UTS} \mathcal{L}_{SI} + \mathcal{L}_{SSI},
    \label{eq:mixture}
\end{equation}
where $\mathbb{I}_{UTS}=1$ for images with UTS or absolute data and $0$ for images with UTSS data. The SSI loss forces the model to generalize and work well for the large-scale UTSS datasets, while the SI loss forces the model to produce unbiased estimates of inverse depth with correct geometry. Since SI loss function (\ref{eq:l1_pairwise}) requires model to predict in log-depth domain and UTSS disparities can not be converted to depth without shift adjustment, our models make predictions in log-disparity domain, which is a negative value of log-depth.

\section{Experimental evaluation}\label{sec:eval}

\paragraph{Network architectures.}

In this work we use a modified Light-Weight Refine Net (LRN) architectures for our experiments. We change the number of channels in CRP and Fusion blocks according to the corresponding output of the backbone in each layer for scalability purposes. As a backbone, as the model optimized for efficiency we use MobileNetV2~\cite{mobilenetv2}, and to increase accuracy we change the backbone to a set of EfficientNet architectures~\cite{efficientnet}. To compare the efficiency, we compute the number of parameters and multiply-addition operations required to infer one sample (in $384\times 384$ resolution). In our approach, models predict the logarithms of inverse depth $\log d^{-1}$. More details of the architecture can be found in the Supplementary.

\paragraph{Implementation details.}

In all experiments, we use the same set of augmentations. We apply resizing, padding, and taking random crops to obtain samples of size 384$\times$384, and then use geometrical transformations (rotation, horizontal flip) and apply color distortions to images (gamma, noise, brightness and contrast). The models were trained using the Ranger optimizer, which is a combination of Radam~\cite{radam} with LookAhead~\cite{lookahead}. We set the learning rate to $10^{-3}$ and use batches of size 32 (except for models based on B3-B5, where we used batches of size 16), with each batch consisting of as equal as possible number of images from each training dataset. We define an epoch as $10{,}000$ training steps and train our models for $40$ epochs. We implemented all models in Python and PyTorch~\cite{pytorch}, using EfficientNets from the \emph{Segmentation Models Pytorch} library~\cite{smp}. Experiments were run with an NVIDIA Tesla P40 GPU. 

\begin{table}[t]
\centering
\begin{tabular}{lcccc}
\hline
        & Scene    & Depth         &  \\
                                Dataset     & type        & type      & \#Samples \\
\hline
\hline
DIML Indoor \cite{diml}         & indoor   & absolute  & 220K  \\
MegaDepth \cite{megadepth}      & general  & UTS       & 130K  \\
ReDWeb \cite{redweb}            & general  & UTSS      & 3600  \\
3D Movies \cite{midas}          & general  & UTSS      & 500K \\
\hline
Sintel \cite{sintel}            & general  & absolute  & 1064 \\
NYUv2 Raw \cite{nyuv2}          & indoor   & absolute  & 407K \\
TUM-RGBD \cite{tumrgbd}         & indoor   & absolute  & 80K \\
DIW \cite{diw}                  & general  & ordinal   & 496K  \\
\hline
\end{tabular}
\caption{Overview of the datasets used in our experiments. \textbf{Top}: training datasets, \textbf{bottom}: test datasets.}
\label{tab:datasets}
\end{table}

\paragraph{Datasets.} Following MIDAS \cite{midas}, we train our models on a mixture of four datasets, specifically MegaDepth \cite{megadepth} ($\approx$100k samples), DIML \cite{diml} ($\approx$220k samples), which are datasets with geometrically complete depth, RedWeb\cite{redweb} (3600 samples), and stereo movies; characteristics of all datasets are summarized in Table~\ref{tab:datasets}. We extend the list of stereo movies used in MIDAS \cite{midas} from 23 to 49. Additionally, we process stereo images with a current state of the art optical flow estimation method, namely RAFT \cite{raft}. This allows us to acquire more accurate disparities with sharp edge boundaries. In total, we use $\approx$500k samples from stereo movies. A comparison of disparity maps and the full list of movies are shown in the Supplementary.

We test our models on several datasets that were completely unseen during training. This test set includes the NYU test set (654 images), the split of the TUM RGBD dataset proposed by Li et al.~\cite{mannequin} (1815 images), \emph{Sintel} dataset (1045 images), and DIW (74441 images). We render dense ground truth depth maps for ETH3D from reconstructed point clouds, similar to~\cite{midas}. Unfortunately, the authors do not provide the resulting depth maps for this dataset, so we have recomputed the metric for their model in our version of ETH3D.

\paragraph{Metrics.}

To evaluate our method, we use standard metrics for depth estimation. To evaluate the results on NYU and TUM datasets, we use the $\delta_{1.25}$ error:
$$
    \delta_{1.25} = \frac{1}{N} \sum_{i=1}^N \mathbb{I}\left[         \max\left\{\frac{{d_i^*}}{d_i}, \frac{{d_i}}{{d_i^*}}\right\} > 1.25\right],
$$
where $ \mathbb{I}\left[ x \right]=1$ if $x$ is true and $0$ otherwise. This metric can be interpreted as the percentage of pixels where the depth deviation from the target exceeds $25\%$.

For Sintel and ETH3D datasets, we use the $\mathrm{rel}$ metric:
$$
    \mathrm{rel} = \frac{1}{M}\sum_{i=1}^M\frac{|{d^*_i} - {d_i}|}{|{d_i^*}|}.
$$
where $d^*_i$ denotes ground truth depth in a pixel $i$ and $d_i$ denotes predicted depth in a pixel $i$. Lower values of this metric indicate better prediction quality.

The DIW dataset contains only one pair of depth ordinal ranking points per image. We evaluate WHDR (Weighted Human Disagreement Rate), i.e., the percentage of incorrectly predicted depth ordinal rankings. Before computing the metrics, we cap maximal depth in datasets to 10, 10, 80, and 72 meters respectively. Before computing UTS metrics for our approach, we align the median of the model's log-disparity prediction to match the ground truth. To compute UTSS metrics, we use an approach similar to MIDAS~\cite{midas}, aligning predictions using MSE criteria. We infer the depth on test images by first resizing their smaller side to $384$ pixels (e.g., $512\times 384$ for the NYU and TUM datasets) and upscaling them back before computing the metrics.

\begin{figure}[t]
    \centering
    \includegraphics[width=0.49\textwidth]{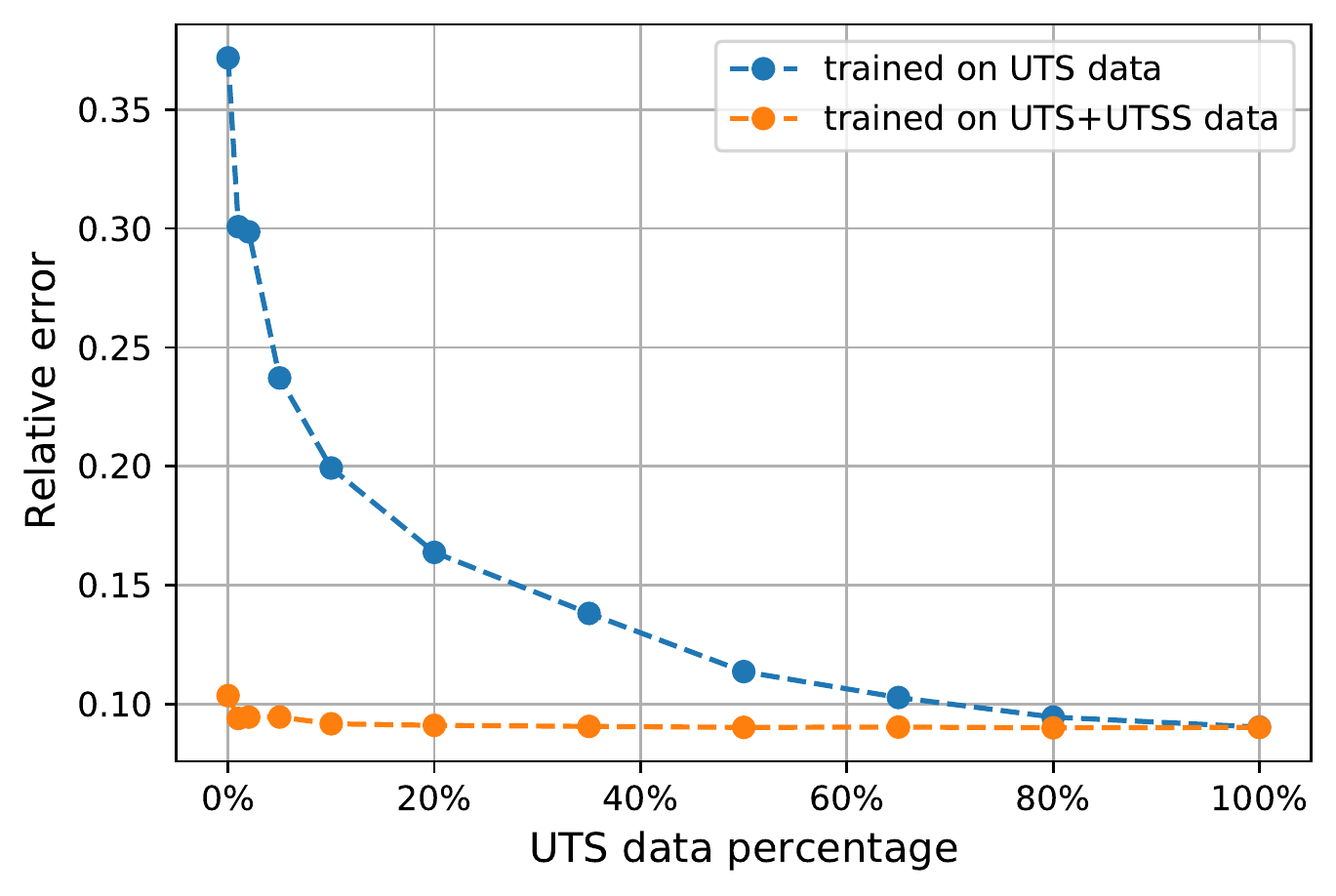}
    \caption{We renormalize depth in the NYU dataset and divide it into two parts: up-to-scale (geometrically complete) and up-to-shift-and-scale (geometrically incomplete). Trained with the proposed method, the model reaches the same quality as the one trained on the full geometrically complete dataset while using only 10-20\% of UTS data. }
    \label{fig:uts_utss}
\end{figure}

\begin{figure*}[!h]
    \centering
    \includegraphics[width=0.195\textwidth]{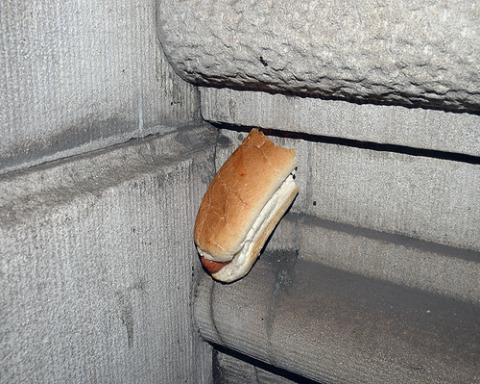}
    \includegraphics[width=0.195\textwidth]{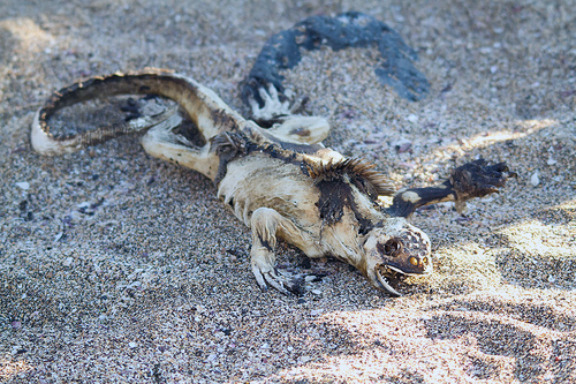}
    \includegraphics[width=0.195\textwidth]{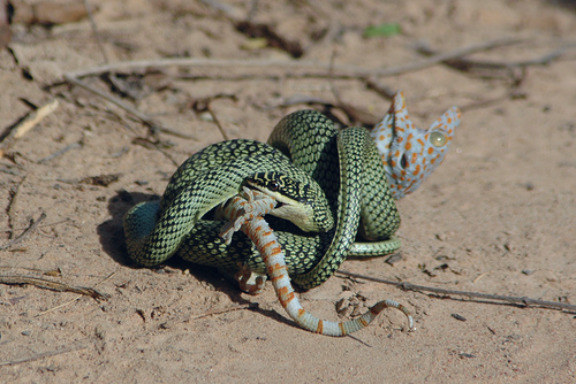}
    \includegraphics[width=0.195\textwidth]{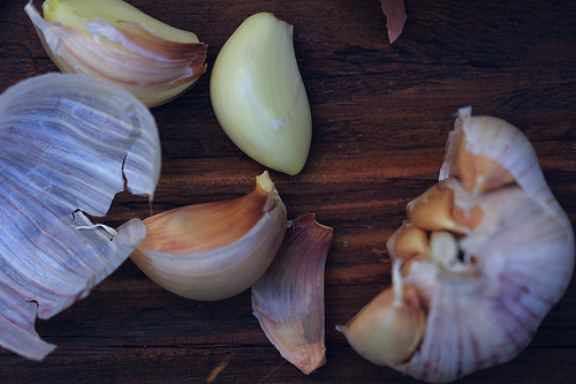}
    \includegraphics[width=0.195\textwidth]{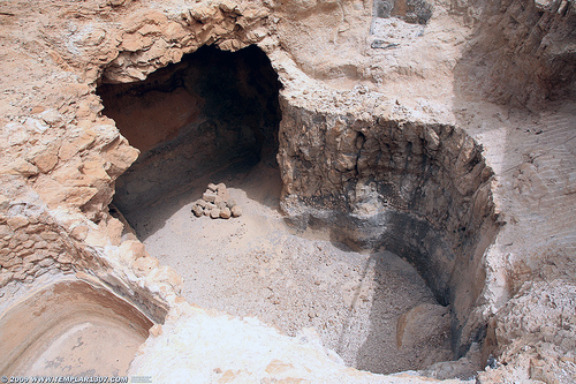} \\
    
    \includegraphics[width=0.195\textwidth]{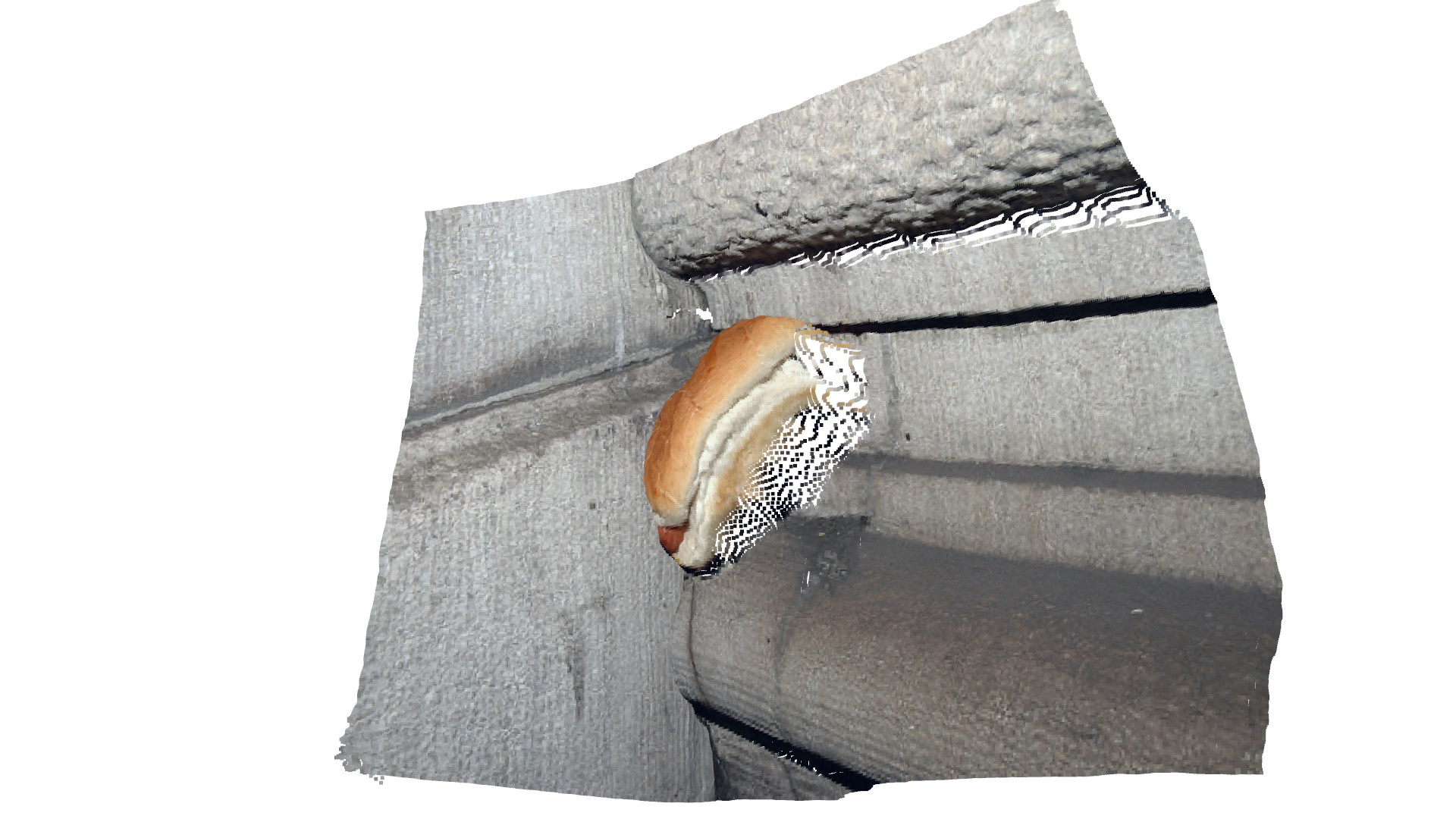}
    \includegraphics[width=0.195\textwidth]{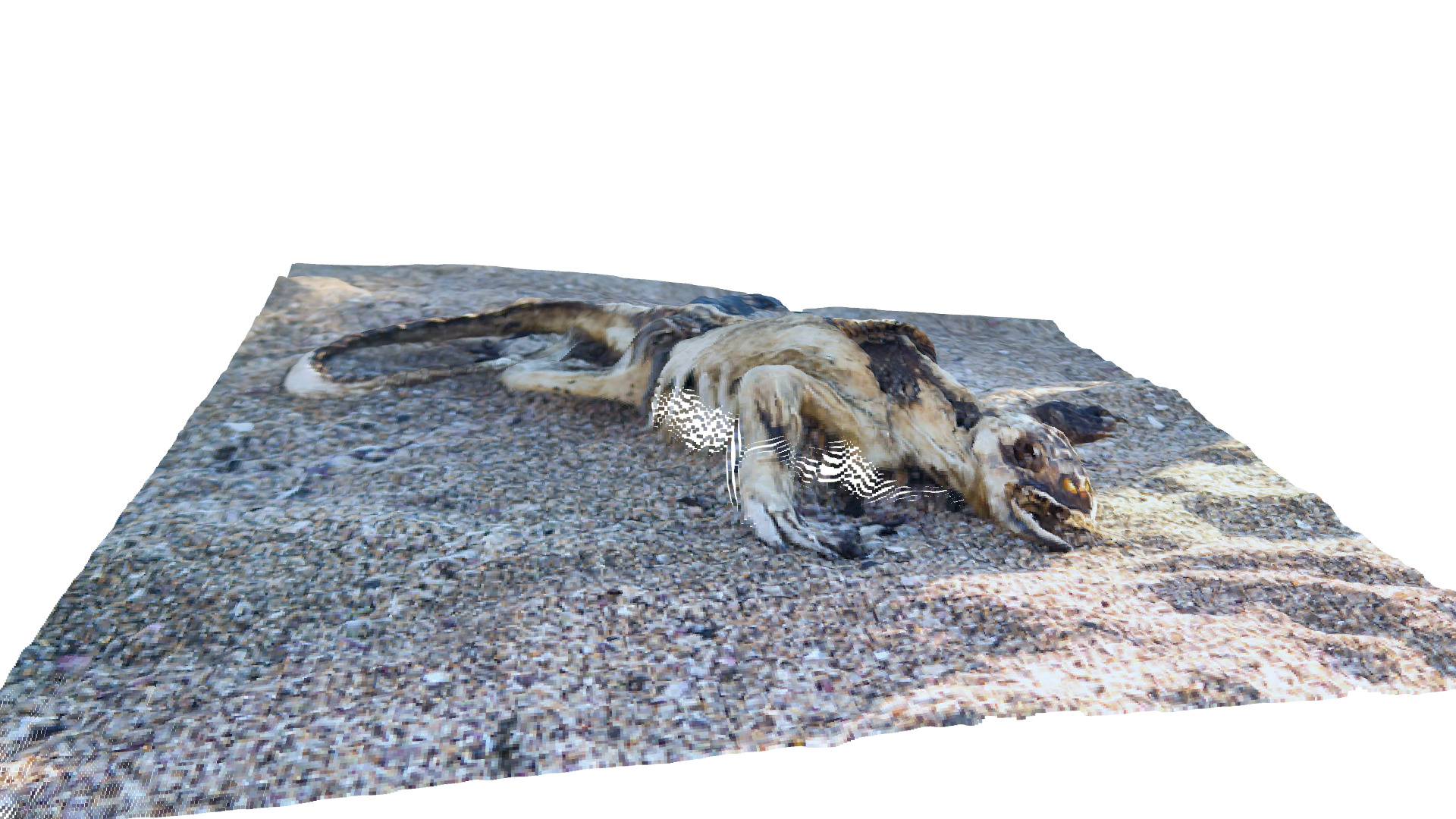}
    \includegraphics[width=0.195\textwidth]{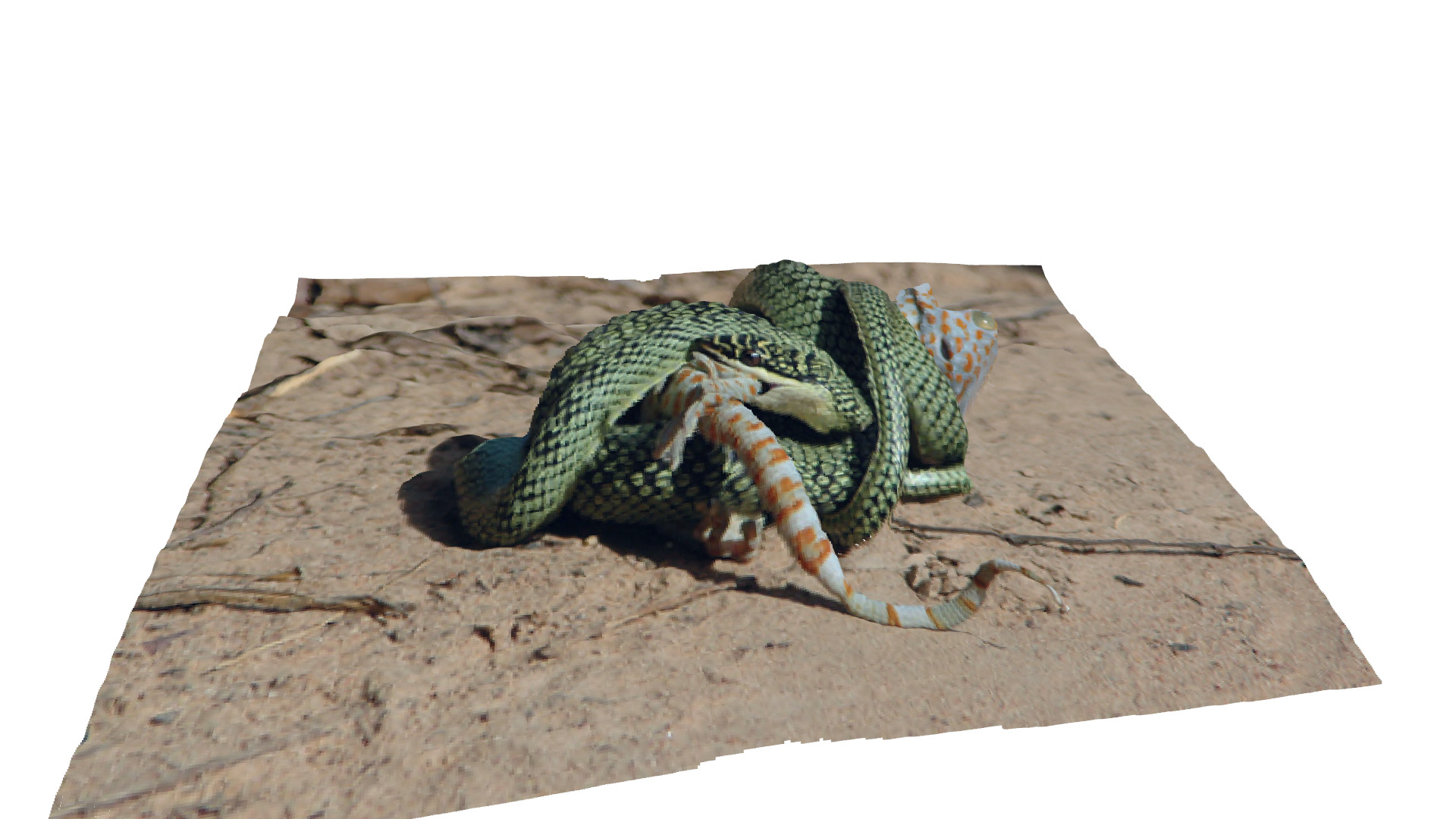}
    \includegraphics[width=0.195\textwidth]{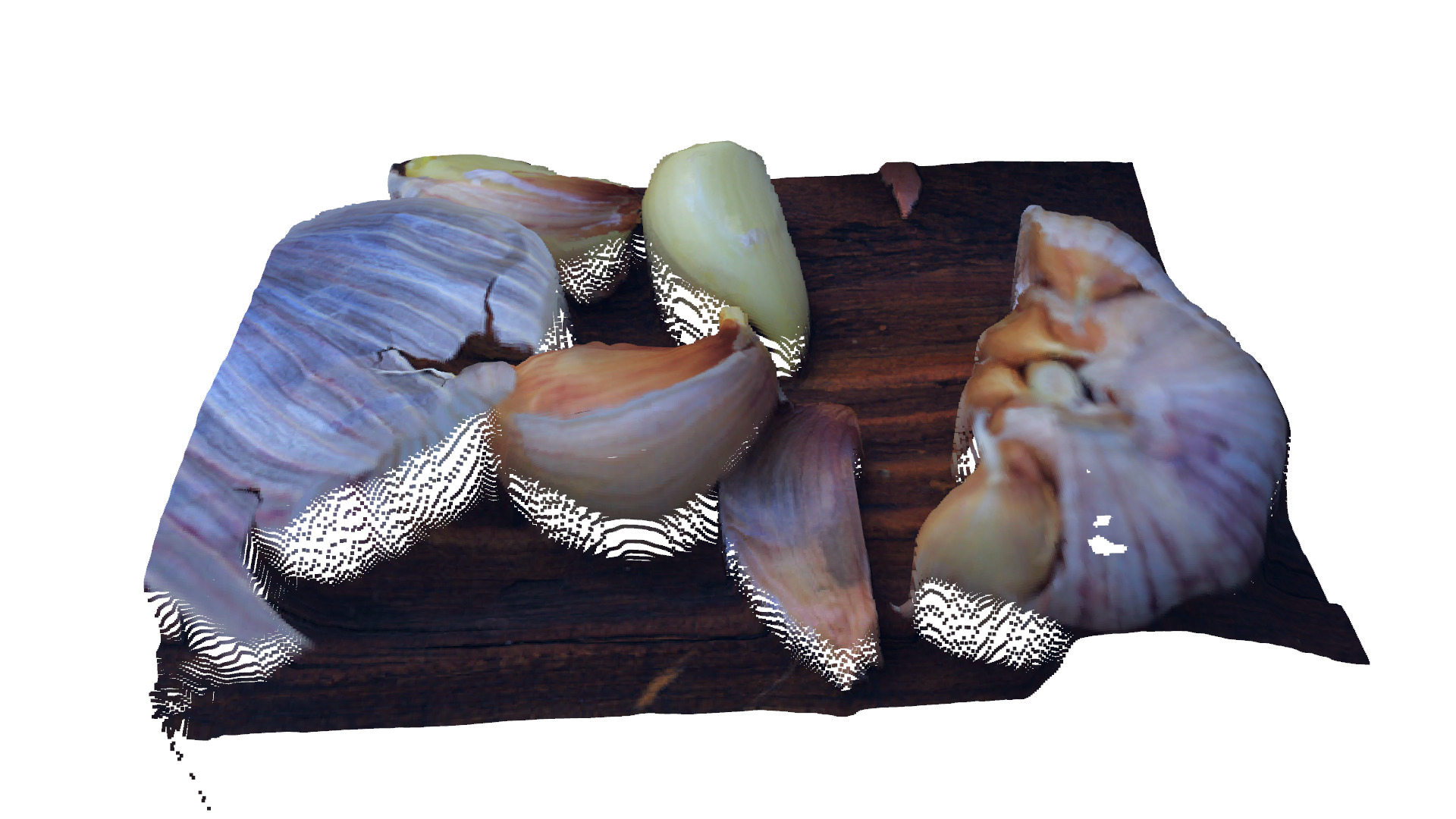}
    \includegraphics[width=0.195\textwidth]{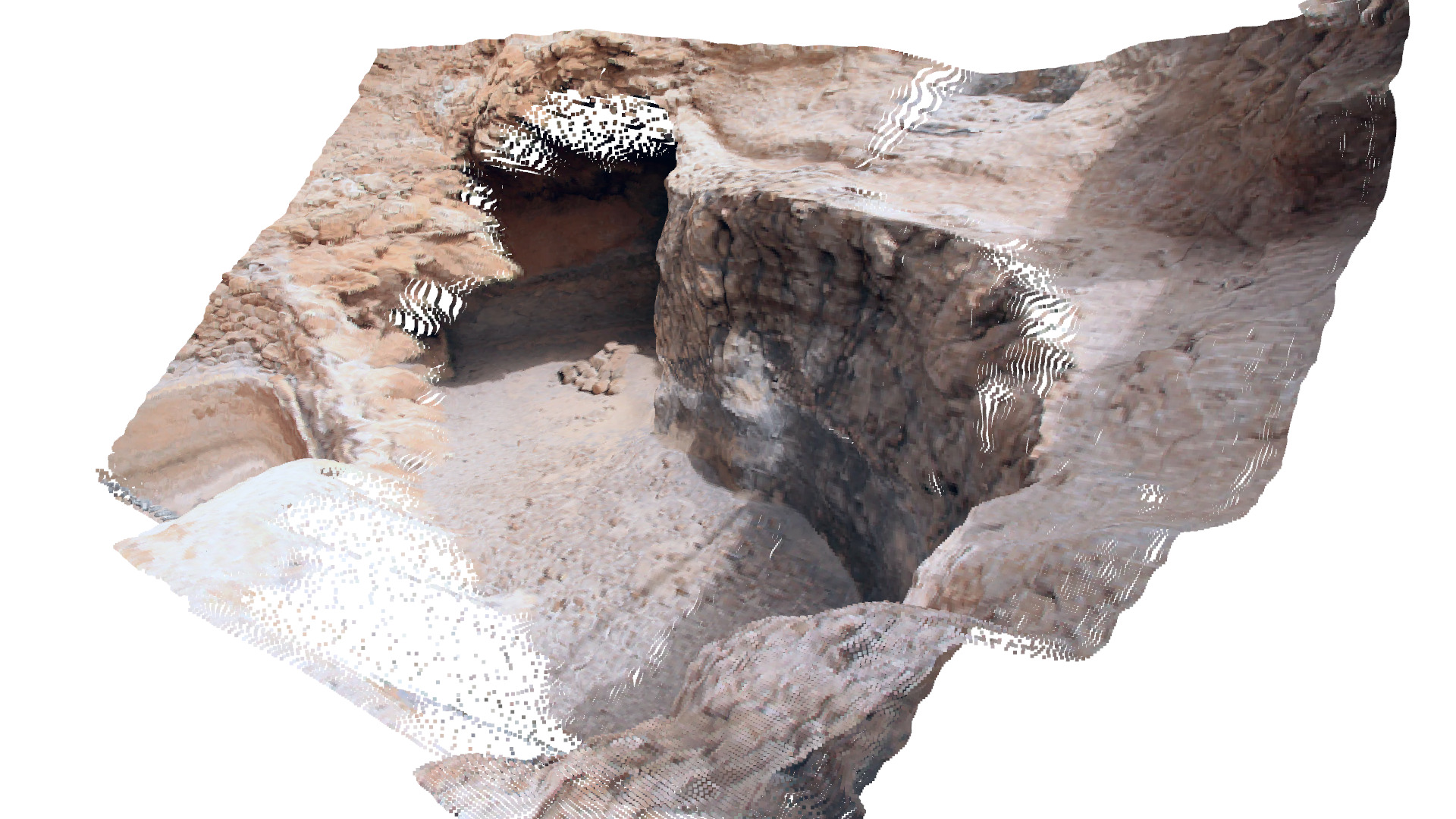}
    \\
    \includegraphics[width=0.195\textwidth]{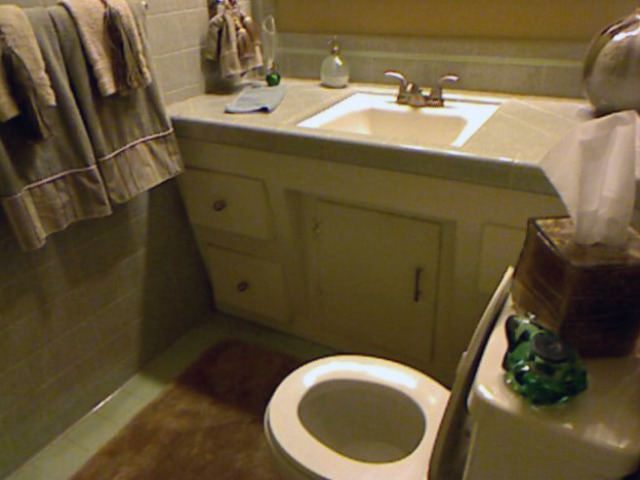}
    \includegraphics[width=0.195\textwidth]{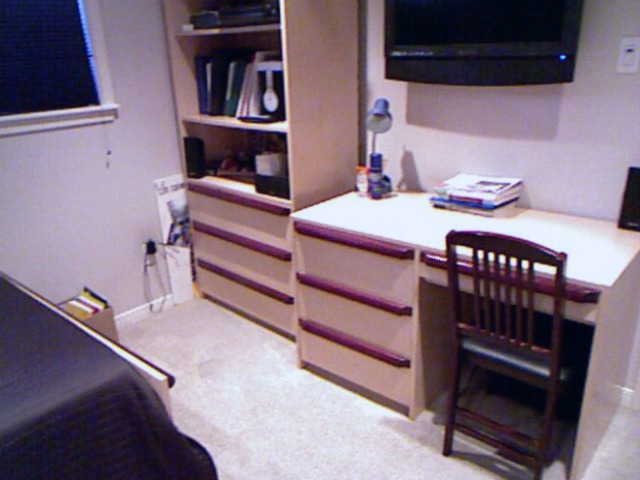}
    \includegraphics[width=0.195\textwidth]{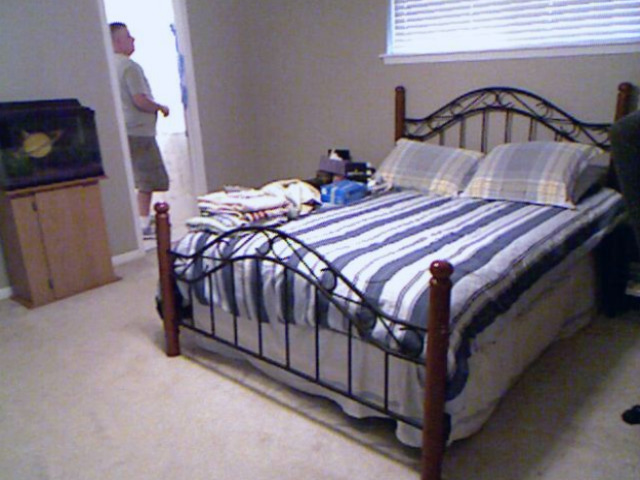}
    \includegraphics[width=0.195\textwidth]{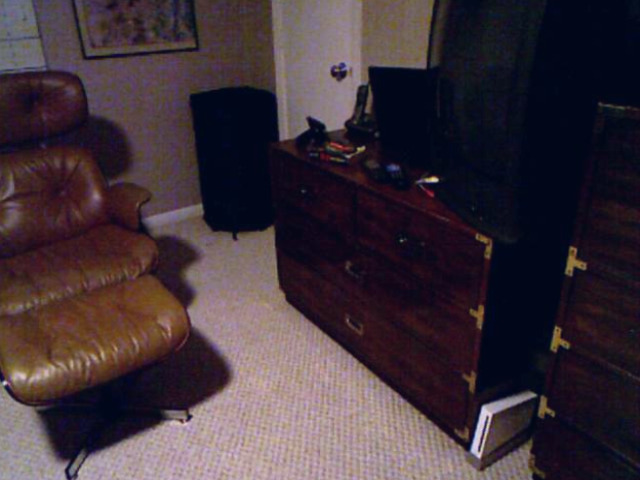}
    \includegraphics[width=0.195\textwidth]{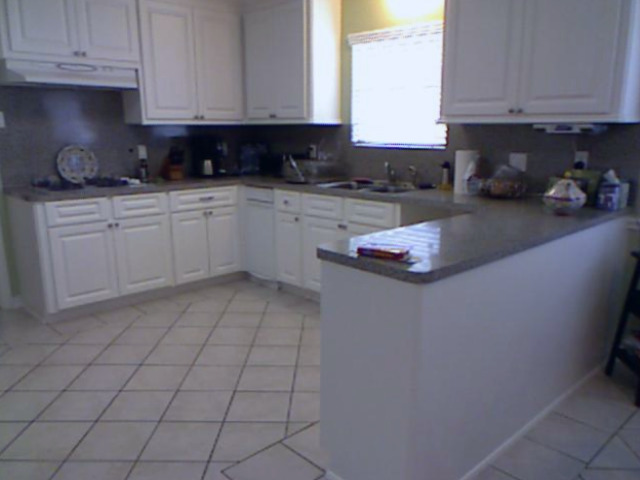}
    \\
    \includegraphics[width=0.195\textwidth]{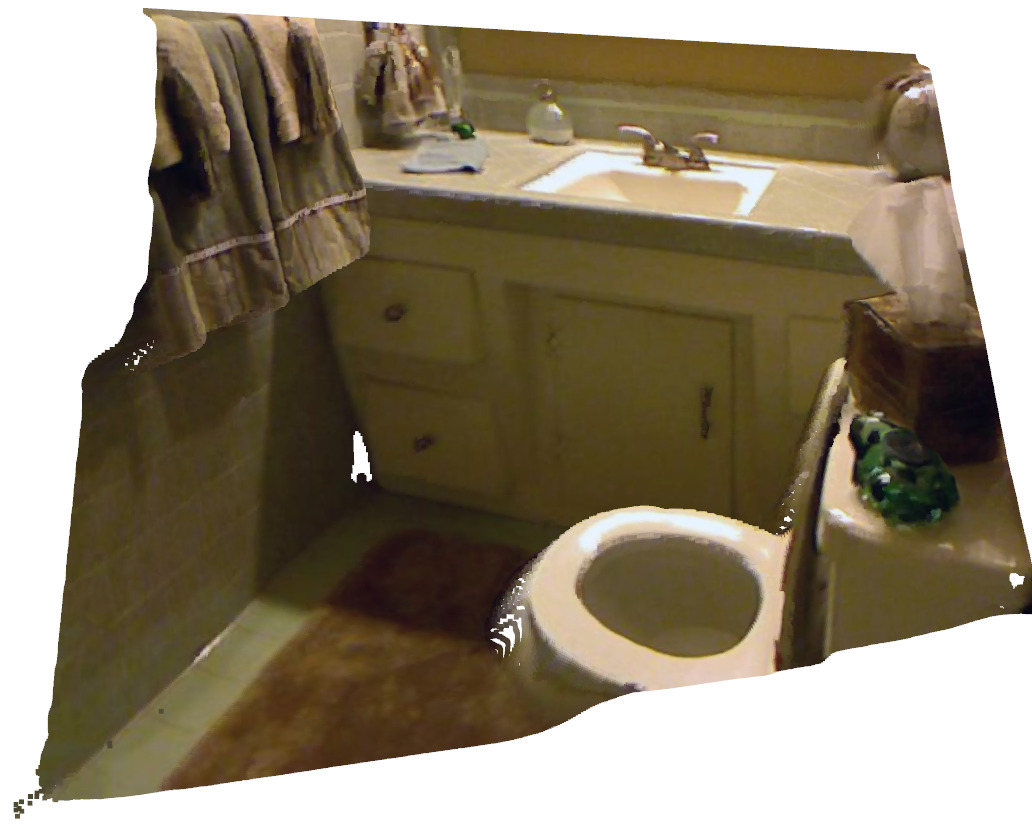}
    \includegraphics[width=0.195\textwidth]{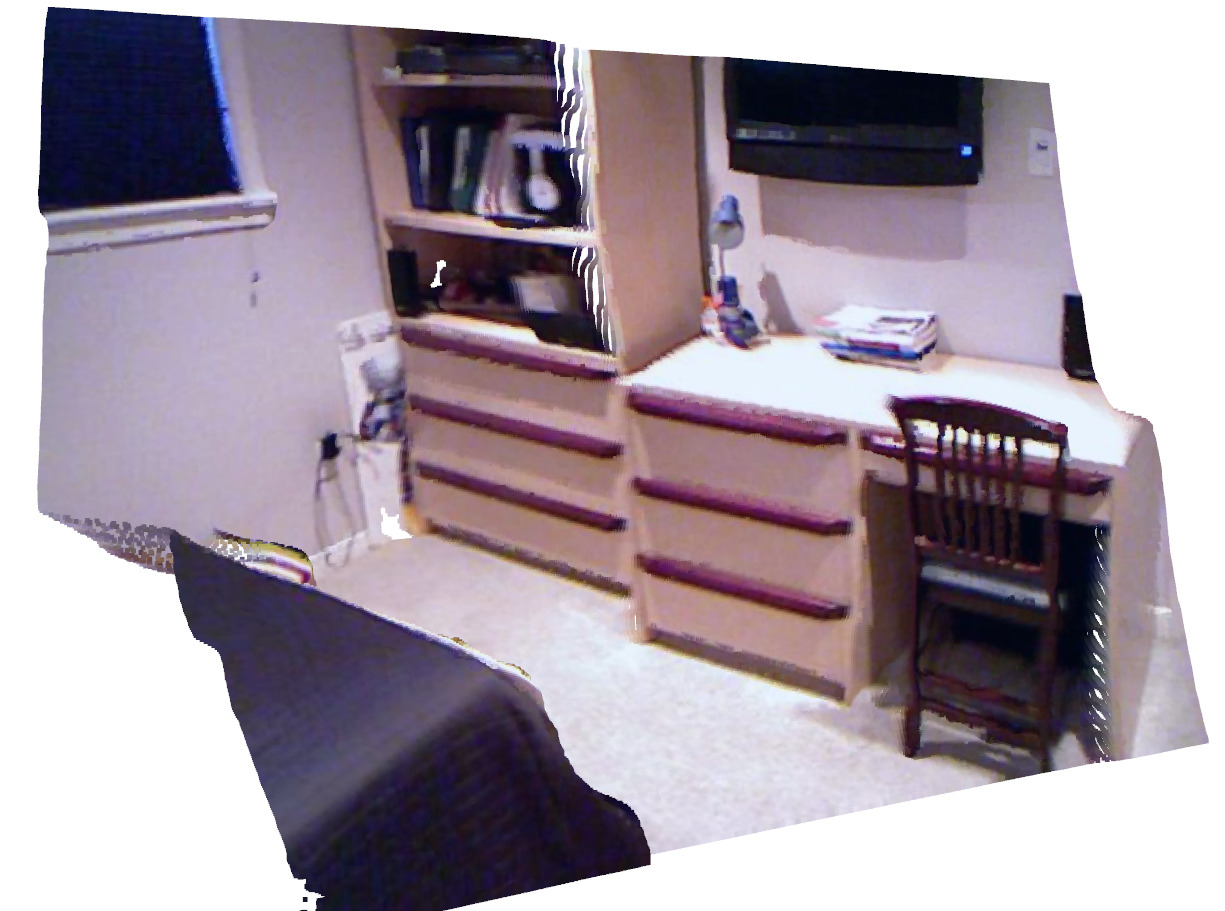}
    \includegraphics[width=0.195\textwidth]{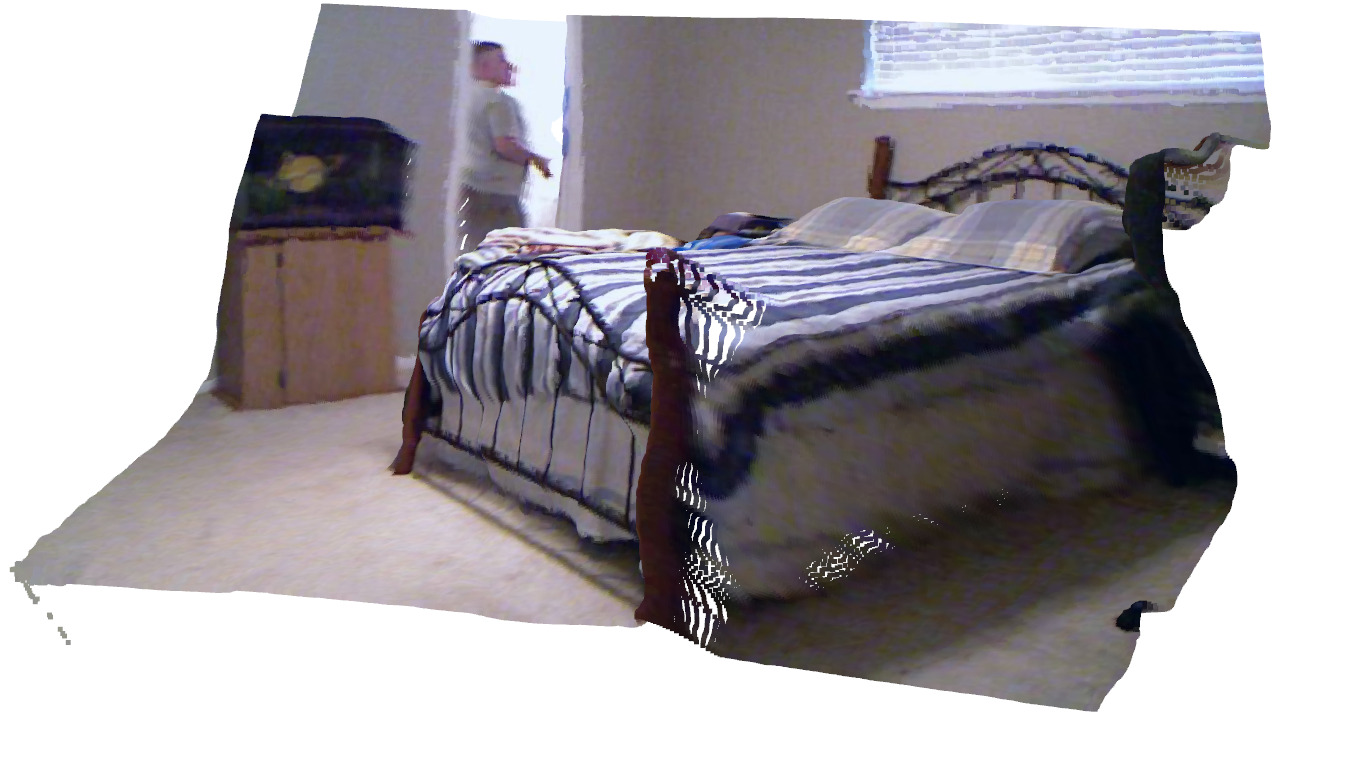}
    \includegraphics[width=0.195\textwidth]{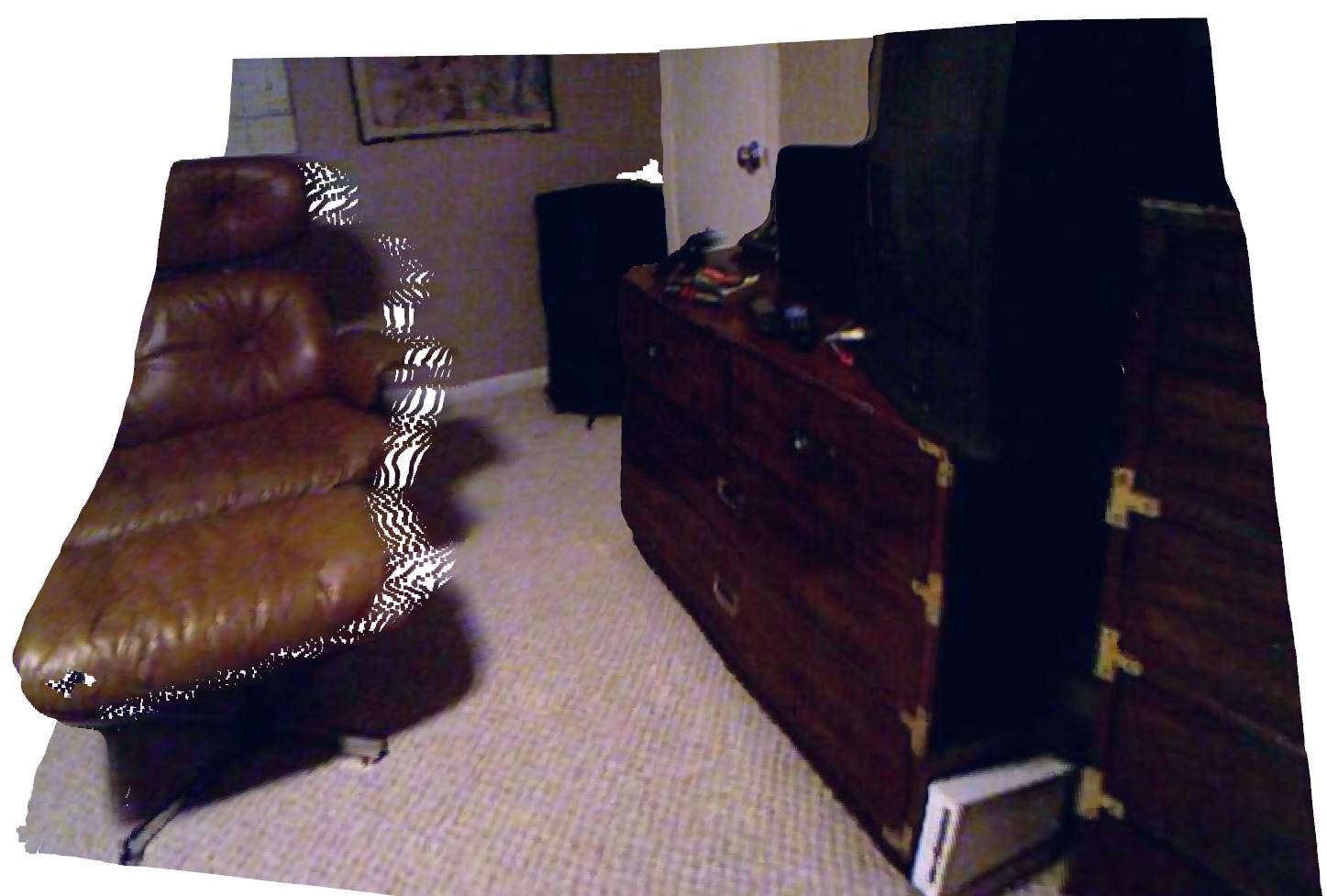}
    \includegraphics[width=0.195\textwidth]{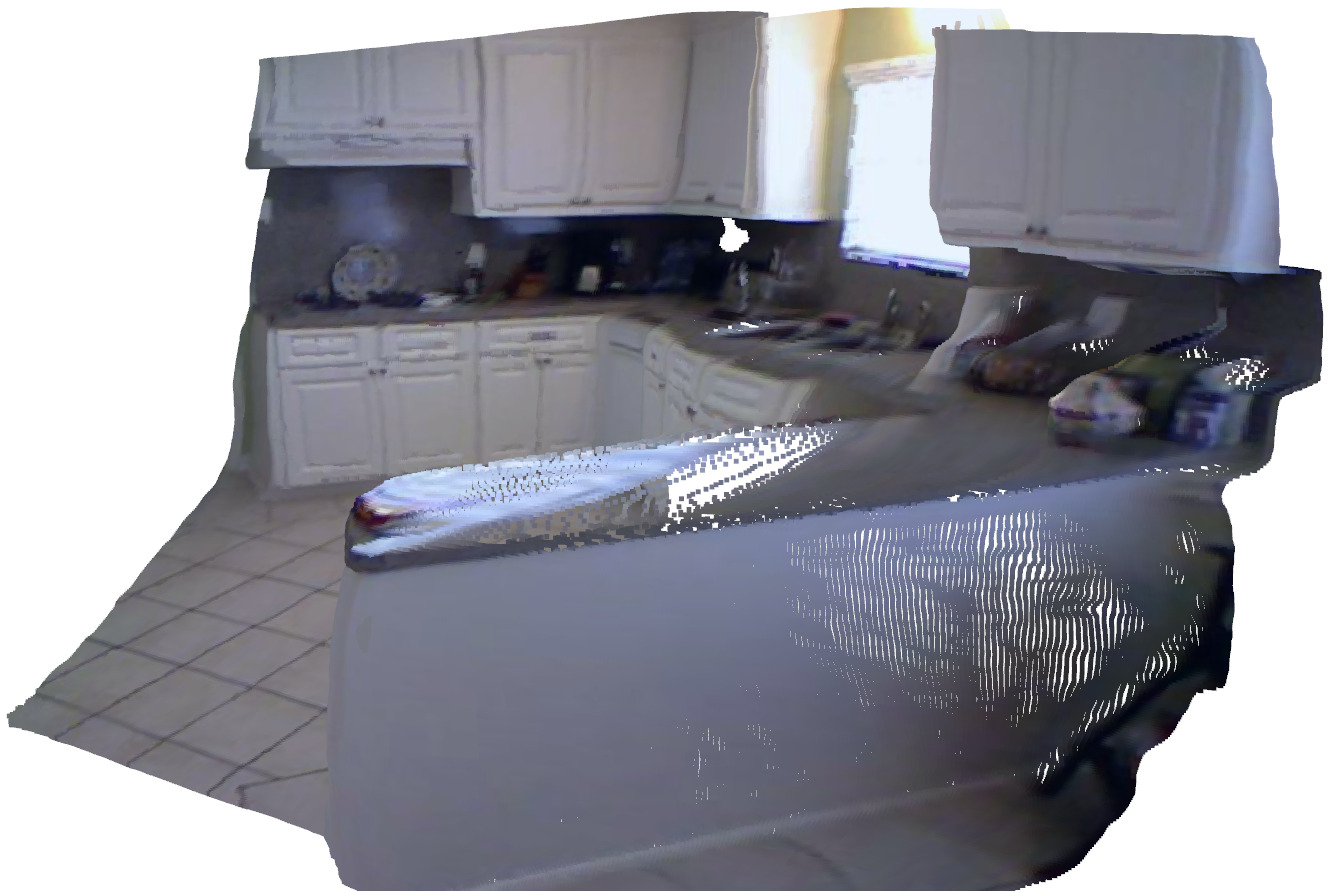}
    
    \caption{Sample images from NYU and DIW datasets (first and third rows) and their corresponding 3D reconstructions (second and fourth rows), obtained with the MN-LRN model. The model was not trained on these datasets yet is able to produce plausible 3D geometry on a wide range of visual scenarios (from indoor to arbitrary outdoor images).}
    \label{fig:meshes}
\end{figure*}

\paragraph*{Ablation study.} 
First, as a proof of concept we train a UTS model on a mixture of UTS and UTSS data from the NYUv2 Raw dataset. Since NYUv2 Raw contains absolute depth, we can convert it to either UTS or UTSS. To convert absolute depth to UTS, we multiply it by a random positive coefficient, and to obtain UTSS data we multiply the inverse depth by a random scale and then shift it by a random additive value.
To obtain comprehensive results, we have tested our approach with several mixtures where UTS and UTSS data are present in different proportions, that is, $p\%$ of the source absolute depth data is converted to UTS data and the rest is converted to UTSS data for a given $p$.

For this proof-of-concept experiment, we use a light-weight model that consists of the MobileNet encoder and LRN decoder. We train it on each UTS/UTSS mixture using the loss function~\eqref{eq:mixture} and then evaluate it against the same model that was trained only on the UTS part of data.

Results of this experiment are shown in Fig.~\ref{fig:uts_utss}. The model trained on UTS data demonstrates expected behaviour: the more training data we use, the better results we obtain. At the same time, the model trained on a mixture of UTS and UTSS data shows similar results for all values of $p$. In other words, it performs as if it was trained on the dataset fully supplied with UTS data.

\begin{table}[t]
\centering\setlength{\tabcolsep}{4pt}
\begin{tabular}{lc|ccc}
\hline
Loss function & Predictions & $\delta_{1.25}$ & rel & $\log_{10}$ \\ \hline\hline
$L_1$-pair & UTS & \textbf{9.27} & \textbf{9.16} & \textbf{0.0388} \\ 
$L_2$-pair~\cite{eigen} & UTS & 9.97 & 9.47 & 0.0402 \\
$L_1$-point & UTS & 9.65 & 9.36 & 0.0397 \\
$L_1$-point & absolute & 10.36 & 9.91 & 0.0419 \\
$L_2$-point & absolute & 11.37 & 10.43 & 0.0446 \\ \hline
\end{tabular}
\caption{Quantitative comparison of commonly used loss functions (data terms) with the proposed $L_1$-pairwise loss on NYU \cite{nyuv2} using MN-LRN model.}
\label{tab:losses}
\end{table}

\begin{table*}[!ht]
\centering
\begin{tabular}{l|ccccc|cc}
\hline
Method                                      & \begin{tabular}{@{}c@{}}NYU \\($\delta_{1.25}$)\end{tabular} & \begin{tabular}{@{}c@{}} TUM \\($\delta_{1.25}$)\end{tabular} & \begin{tabular}{@{}c@{}}ETH3D \\($\mathrm{rel}$)\end{tabular} & \begin{tabular}{@{}c@{}}Sintel \\($\mathrm{rel}$)\end{tabular}& \begin{tabular}{@{}c@{}}DIW \\(WHDR) \end{tabular}&\begin{tabular}{@{}c@{}}Params \\(mln) \end{tabular}       & \begin{tabular}{@{}c@{}}MAdds \\($10^9$) \end{tabular}        \\ \hline\hline
Li et al. \cite{megadepth} & 34,39         & 33,11          &   0.276      & 0.490          & 24,55          & 5.4          & 91.1               \\
Mannequin \cite{mannequin} & 23.42         & 22,39          &   0.249       & 0.431          & 26.52          & 5.4          & 91.1                \\ 
Tiefenrausch \cite{kopf3d}       & 32.8          & 35.4           &  0.314          &  0.497 &  25.54 & 3.6 & 7  \\
\hline
MN-LRN                                      & 14.64         & 15.13          &   0.191        & 0.360          & 15.02          & \textbf{2.4} & \textbf{1.17} \\
EfficientNet-Lite0-LRN                      & 14.15         & 14.41          &  0.177         & 0.354          & 14.59          & 3.6          & 1.29                   \\
EfficientNet-B0-LRN                         & 13.84         & 15.95          &  0.168         & 0.330          & 13.15          & 4.2          & 1.66                 \\
EfficientNet-B1-LRN                         & 12.80         & 15.03          &   0.179        & 0.315          & 12.71    & 6.7          & 2.22              \\
EfficientNet-B2-LRN                         & 13.04         & 15.36          &   0.168        & \textbf{0.304} & 13.06          & 8            & 2.5               \\
EfficientNet-B3-LRN                         & 12.35         & 14.38          &   0.176        & 0.343          & 12.95          & 11           & 3.61                 \\
EfficientNet-B4-LRN                         & 11.92         & {13.55}        &   0.164        & 0.346          & 12.81          & 18           & 5.44            \\
EfficientNet-B5-LRN                         & \textbf {10.64}   & \textbf{13.05} &  \textbf{0.154}  & 0.328 & \textbf{12.56}         & 29           & 8.07        \\ \hline
\end{tabular}
\caption{Results of the UTS models trained on the datasets mixtures of UTS and UTSS data compared to other UTS single-view depth estimation methods.}
\label{tab:light_weight}
\end{table*}

\begin{figure*}[!h]
\centering
\begin{tabular}{cc|ccccc}
\hline
Model                  & Data     & NYU            & TUM            & ETH3D         & Sintel         & DIW            \\ \hline\hline
MN-LRN                 & UTS      &   15.38        &  17.78        &    0.193       &    0.432       & 21.49          \\
MN-LRN                 & UTS+UTSS & \textbf{14.64} & \textbf{15.13} & \textbf{0.191} & \textbf{0.360} & \textbf{15.02} \\ \hline
EfficientNet-Lite0-LRN & UTS      & \textbf{14.15} & 16.99         &    0.191      & 0.428          & 19.70          \\
EfficientNet-Lite0-LRN & UTS+UTSS & \textbf{14.15} & \textbf{14.41} & \textbf{0.177} & \textbf{0.354} & \textbf{14.59} \\ \hline
\end{tabular}
\captionof{table}{Results of the models trained on UTS and combined UTS and UTSS data. For both models, UTS+UTSS data combination significantly improves prediction quality. At the same time, the models retain the ability to compensate for $C_2$ since only scale-invariant metrics were used.}
\label{tab:uts_utss}\vspace{.5cm}


\centering
\begin{tabular}{l|ccccc|cc}
\hline
Model name &  NYU & TUM & ETH3D & Sintel & DIW  & Params & FLOPS \\ \hline \hline
MIDAS \cite{midas} & 9.55 & 14.29 & 0.167 & 0.327 & \textbf{12.46}  & 105.4 & 103.9 \\
Ours, MN-LRN & 10.97 & 14.22 & 0.177 & 0.292 & 15.02 & 2.4 & 1.17 \\
Ours, B5-LRN & \textbf{7.4} & \textbf{9.86} & \textbf{0.145} & \textbf{0.253} & 12.56  & 29 & 8.07 \\
\hline
\end{tabular}
\caption{Comparison of the proposed models with the MiDaS model~\cite{midas}. For this comparison, UTSS metrics were used.}
\label{tab:midas-comp}
\vspace{.3cm}

    \centering
    \includegraphics[width=0.98\textwidth]{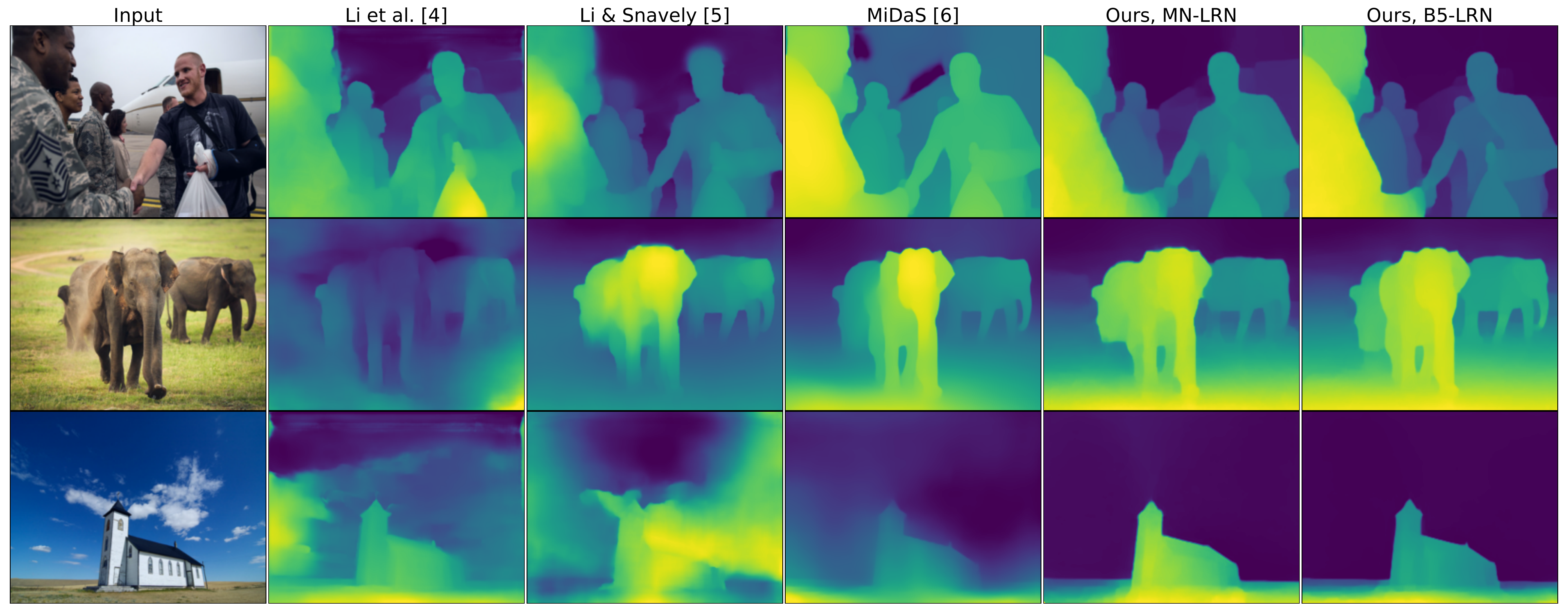}
    \captionof{figure}{Qualitative comparison of depth maps produced by our models and existing competitors. Images are taken from the DIW dataset and were not seen during training. }
    \label{fig:midas_results}
\end{figure*}

\paragraph*{Loss functions.}

We have conducted a set of experiments on the NYUv2 dataset in order to find the best out of the loss functions that we will use in large-scale training. Experimental results are presented in Table~\ref{tab:losses}. These experiments have shown that our new loss function yields better results on all metrics.

This proof of concept experiment demonstrates that a lack of geometrically correct data can be compensated for by using extra UTSS data. Moreover, the proportion of UTS and UTSS data does not affect the final result much. This means that one can use a relatively small amount of geometrically correct data (provided, e.g., in NYUv2 or MegaDepth dataset) and accompany it with large-scale, diverse UTSS datasets that can be found, e.g., in stereo movies in virtually unlimited amounts. This experiment suggests that a model trained with this mixture might behave as well as a model trained on a similar amount of data \emph{all} supplied with geometrically correct depth maps.

Along with an ablation study of the loss functions that can be found in Table~\ref{tab:losses}, we also perform yet another proof-of-concept experiment. We compare two identical models, one trained to predict the absolute depth map ($L_1$-pointwise, absolute) and the other trained to predict the UTS depth map ($L_1$-pointwise, UTS). The latter model achieves better values of UTS metrics. 

\paragraph{Training on a mixture of datasets}

We train our networks on several datasets at the same time. Among training datasets used in this experiment, only MegaDepth and DIML contain UTS data (see also a full comparison in Table~\ref{tab:datasets}). Consequently, we can train on these two datasets in UTS mode and on all the training datasets in UTSS mode. In Table~\ref{tab:light_weight} we compare our models with commonly used popular depth estimation models using UTS metrics. We also compare our solution with the results of~\cite{midas} in Table~\ref{tab:midas-comp}. In Table~\ref{tab:uts_utss} we compare the performance of models with and without UTSS data.

In Fig.~\ref{fig:pointclouds}, we compare point clouds that are generated using our B5-LRN model with point clouds built using other popular models. We also show some point clouds generated by our most efficient network MN-LRN in Fig.~\ref{fig:meshes}. Note that the images that we are using in these demonstrations are taken from datasets unseen during training, showcasing the generalization capabilities of our models. The point clouds produced using our method look competitive compared to the other methods.
More visualizations of the results are provided in the Supplementary.


\paragraph{Discussion.}

Acquisition of UTS depth data is often a bottleneck. At the same time, sources of UTSS stereo data are accessible, diverse, and plentiful.
In our evaluation study, we have seen that we can train a UTS model better by supplementing the dataset with UTSS data. This broadens the horizons for production-ready solutions. We argue that stereo data gathered from multiple sources together with existing UTS and absolute depth datasets provide a solid basis for a versatile and robust SVDE method.
Obtaining depth maps via stereo matching has certain limitations. First, precise disparities for distant objects can be obtained only from a very wide stereo baseline. Second, the method depends massively on the model used to estimate the disparity. To estimate depth in the distance, it should be sensitive enough to capture small displacements. Accordingly, SVDE is not yet applicable for large-scale scenarios such as outdoor landscapes.

Furthermore, our results indicate that lightweight models are not yet able to produce sharp depth maps. Still, there is a trade-off between the accuracy of depth estimates and the complexity of the model that determines speed of the inference, memory requirements, and power consumption.

\section{Conclusion}\label{sec:concl}

In this work, we have shown that training geometry-preserving SVDE models benefits significantly from the use of voluminous data from stereo pairs. The resulting models exhibit strong generalization capabilities and produce robust depth maps that can be used to generate natural point clouds from a single input image. Although the results of these models are still imperfect, they already yield quality sufficient for many practical applications.

We have also presented a family of models that produce state of the art results for geometrically preserving depth estimation on the majority of existing datasets. One of these models is light-weight, computationally efficient, and based on a mobile-oriented backbone architecture. This enables general purpose SVDE to be used on consumer devices.

\FloatBarrier

{\small
\bibliographystyle{ieee_fullname}
\bibliography{main}
}

\newpage

\appendix
\section{Depth uncertainty impact on geometry estimation}
\label{apdx:utss_depth}

For several datasets, sourced from stereo films \cite{midas} and arbitrary stereo photos from the internet \cite{redweb}, ground-truth disparity can be obtained only up to unknown scale and shift coefficients. Without knowing the correct disparity shift value scene 3D geometry can not be reconstructed properly. To illustrate this, suppose a 3D line that is not aligned with the optical axis of a pinhole camera. Depth for the 3D line projection point $x, y$ can be expressed as:

\begin{equation}
    d = ax + by + c,
\end{equation}

where $a, b, c$ are some coefficients and $x, y$ belongs to the point set on the camera matrix containing the 3D line projection. Suppose that an inverse depth (disparity) of this 3D line is defined up to unknown shift and scale:

\begin{equation}
    \frac{1}{\tilde{d}} = \frac{C_1}{ax + by + c} + C_2,
\end{equation}

or equivalently

\begin{equation}
    \tilde{d} = \frac{ax + by + c}{C_1 + C_2 (ax + by + c)}.
\end{equation}

This expression denotes a line iff $C_2$ is zero. Therefore, to obtain predictions useful for 3D scene reconstruction, a neural network should evaluate the $C_2$ coefficient explicitly.

Though the $C_2$ coefficient significantly affects the point cloud's geometry, $C_1$ affects only the global scene scale. To illustrate that, we can consider mapping from the pinhole camera plane point $(x,y)$ and its corresponding depth $d$ to the 3D scene point:

\begin{equation}
\begin{pmatrix}x  \\y \\ d \end{pmatrix} \mapsto \begin{pmatrix} \frac{(x-c_x)d}{f_x} \\ \frac{(y-c_y) d}{f_y} \\ d\end{pmatrix}.
\label{eq:pinhole}
\end{equation}

Suppose the original depth map is scaled by a factor $C_1$. According to \ref{eq:pinhole}, all the 3D point coordinates are also multiplied by $C_1$. Thus the overall scene is just scaled by $C_1$ without affecting the correctness of the geometry (e.g. angles and curvatures). 

\section{Efficient computation of the proposed loss function}
\label{apdx:l1_pairwise}

Inspired by the loss function of Eigen et al. \cite{eigen}, we propose up-to-scale pairwise $L_1$ loss function:
\begin{equation}
    L = \frac{1}{N^2} \sum_{i,j=1}^N | (\log d_i - \log d_j) - (\log d_i^* - \log d_j^*) |.
\end{equation}

This loss function is scale-invariant since the difference of log-depth values eliminates unknown depth scale coefficients. 

As the original expression requires to sum up $O(N^2)$ terms, we propose an efficient way of computing it. Using a substitution $R_i = \log d_i - \log d_i^*$ we get:
\begin{equation}
    L = \frac{1}{N^2} \sum_{i, j=1}^N | R_i - R_j|,
\end{equation}
or equivalently:
\begin{equation}
    L = \frac{1}{N^2} \sum_{i,j=1}^N | R_{\{i\}} - R_{\{j\}} |,
\end{equation}

where $R_{\{i\}}$ is an $i$-th element in ascending order. By changing the order of summation, one can get the following expression:
\begin{equation}
    L = \frac{2}{N^2} \sum_{j = 1}^N \sum_{i = j + 1}^N \left( R_{\{i\}} - R_{\{j\}}\right)
\end{equation}

since the expression in the brackets is always positive. In this summation, term $R_{i}$ occurs $i - 1$ times with a positive sign and $N - i$ times with a negative sign. Hence, sum of these terms is equal to $(2i - N - 1) R_{\{i\}}$. This can be rewritten as
\begin{equation}
    L = -\frac{2}{N^2} \sum_{i=1}^N (N - 1 - 2 (i - 1)) R_{\{i\}}.
\end{equation}
Thus, $L_1$ pairwise loss can be computed in $O(N\log N)$ time, which is a complexity of sorting operation. However, in practice, we faced only negligible training time increase compared to the training time with conventional loss functions.

\section{Overview of alternative loss functions}

There are several other functions that can be used for geometry preserving SVDE task. First of all there are absolute loss functions that are designed for absolute depth prediction. First one may be called $L_2$ pointwise loss function:
\begin{equation}
    L = \frac{1}{N} \sum_{i = 1}^N (\log d_i - \log d_i^*)^2.
\end{equation}

In this loss function the discrepancy between the logarithms of target and predicted depths is penalized in each of the pixels.

Secondly, there is an $L_1$ pointwise loss function that is known to be more robust to outliers:
\begin{equation}
    L = \frac{1}{N} \sum_{i=1}^N |\log d_i - \log d_i^*|.
\end{equation}

Both loss functions, $L_1$ pointwise and $L_2$ pointwise, can be modified to become scale-invariant and to be able to work with UTS data. $L_2$ pointwise loss function should me modified as follows:
\begin{equation}
    L = \frac{1}{N} \sum_{i=1}^N (\log d_i - \log d_i^* - \mu)^2,
\end{equation}

where

\begin{equation}
    \mu = \frac{1}{N} \sum_{i=1}^N (\log d_i - \log d_i^*).
\end{equation}

$L_1$ pointwise loss function can be modified similarly to become a scale-invariant loss, the only difference is that instead of mean value $\mu$, the median value should be used.

Finally, there is $L_2$ pairwise loss function \cite{eigen2015predicting}:
\begin{equation}
    L = \frac{1}{N^2} \sum_{i=1}^N \left((\log d_i - \log d_j) - (\log d_i^* - \log d_j^*)\right).
\end{equation}

This loss function may be computed in $O(N)$ time using the formula:

\begin{equation}
    L =  \frac{1}{N} \sum_{i=1}^N R_i^2 - \frac{1}{N^2} \left(\sum_{i=1}^N R_i\right)^2.
\end{equation}

$L_2$ pairwise loss function is scale-invariant and, thus, can be used for training on UTS data. It compares all the pairs of the pixels on the image.

\section{Architecture modifications}
\label{apdx:architecture}

\begin{figure}\centering
    \includegraphics[width=0.45\textwidth]{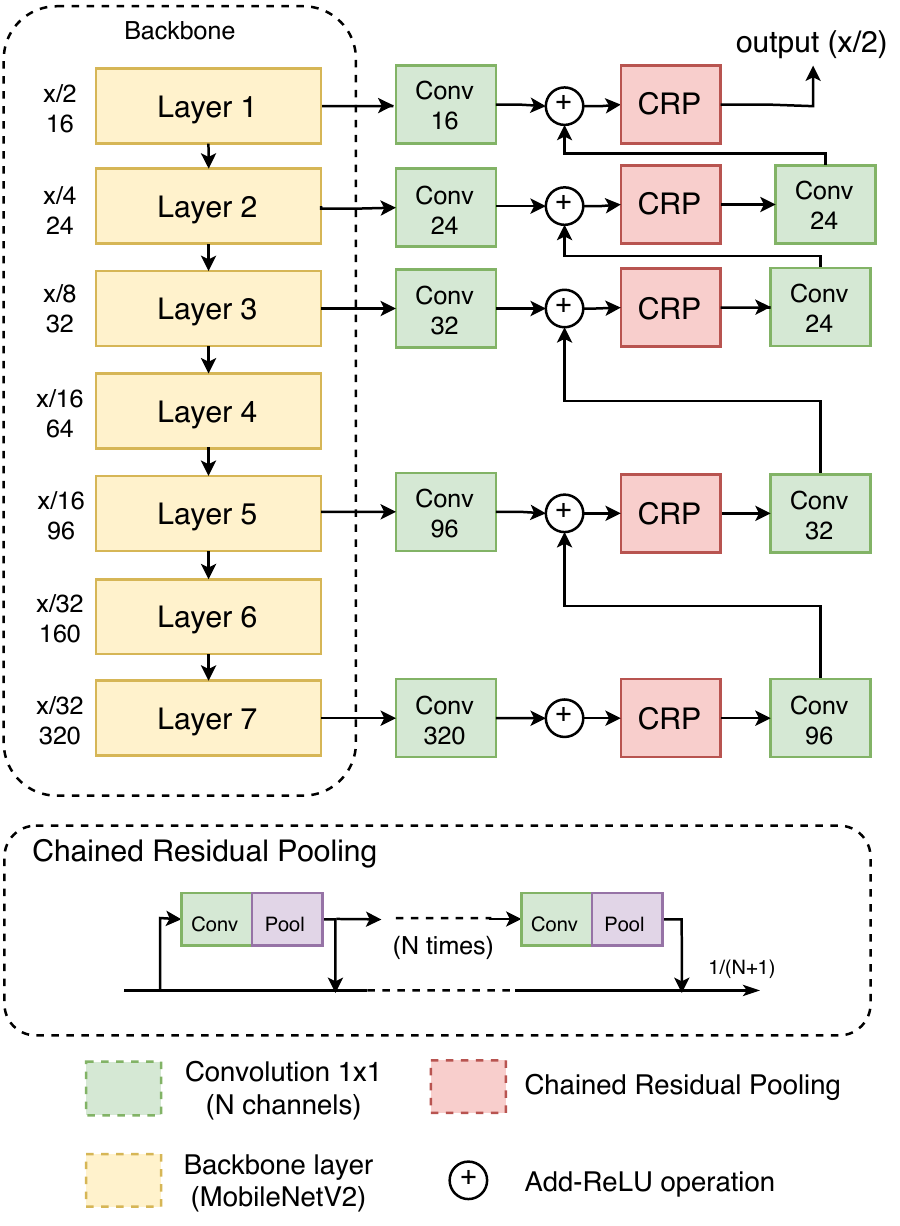}
    \caption{\textbf{Top:} Architecture used in our method. The number of channels in fusion convolutions is equal to the number of channels in the corresponding backbone level to provide decoder scalability. \textbf{Bottom:} updated CRP block. If the CRP block has $N$ CRP modules, the output signal should be divided by $N + 1$.}
    \label{fig:lrn}
\end{figure}

\begin{table}[h!]
    \centering
\begin{tabular}{|l|r|r|}
\hline
Name                                        & Year & Frames \\
\hline
\hline
3-D Sex and Zen: Extreme Ecstasy            & 2011                     & 12201                      \\ \hline
\begin{tabular}[c]{@{}l@{}}A Very Harold \& \\ Kumar 3D Christmas\end{tabular} & 2011                     & 6418                       \\ \hline
Battle of the Year                          & 2013                     & 10992                      \\ \hline
Cirque du Soleil: Journey of Man            & 2000                     & 3654                       \\ \hline
Creature from the Black Lagoon              & 1954                     & 7266                       \\ \hline
Dark Country                                & 2009                     & 7657                       \\ \hline
Dolphin Tale                                & 2011                     & 11536                      \\ \hline
 & 2009                     & 3077                       \\ \hline
Drive Angry                                 & 2011                     & 10679                      \\ \hline
Exodus: Gods and Kings                      & 2014                     & 14855                      \\ \hline
Final Destination 5                         & 2011                     & 9009                       \\ \hline
Flying Swords of Dragon Gate                & 2011                     & 13301                      \\ \hline
Galapagos: The Enchanted Voyage             & 1999                     & 1787                       \\ \hline
Ghosts of the Abyss                         & 2003                     & 6300                       \\ \hline
Hugo                                        & 2011                     & 12852                      \\ \hline
Into the Deep                               & 1994                     & 1564                       \\ \hline
Jack the Giant Slayer                       & 2013                     & 10020                      \\ \hline
Journey 2: The Mysterious Island            & 2012                     & 9923                       \\ \hline
Journey to the Center of the Earth          & 2008                     & 9472                       \\ \hline
Life of Pi                                  & 2012                     & 9926                       \\ \hline
My Bloody Valentine                         & 2009                     & 10275                      \\ \hline
Oz the Great and Powerful                   & 2013                     & 11087                      \\ \hline
Pina                                        & 2011                     & 10674                      \\ \hline
Piranha 3DD                                 & 2012                     & 7718                       \\ \hline
\begin{tabular}[c]{@{}l@{}}Pirates of the Caribbean: \\ On Stranger Tides\end{tabular} & 2011                     & 12914                      \\ \hline
Pompeii                                     & 2014                     & 9178                       \\ \hline
Prometheus                                  & 2012                     & 11114                      \\ \hline
Sanctum                                     & 2011                     & 9682                       \\ \hline
Saw 3D: The Final Chapter                   & 2010                     & 8632                       \\ \hline
\begin{tabular}[c]{@{}l@{}}Sea Rex 3D: \\ Journey to a Prehistoric World\end{tabular} & 2011                     & 4130                       \\ \hline
\hline
\end{tabular}
    \caption{Stereo movies used in our experiments, part 1}
    \label{tab:3dmv-list1}
\end{table}

\begin{table}[h!]
    \centering
\begin{tabular}{|l|r|r|}
\hline
Name                                        & Year & Frames \\ \hline\hline
Silent Hill: Revelation 3D                  & 2012                     & 8533                       \\ \hline
Sin City: A Dame to Kill For                & 2014                     & 10788                      \\ \hline
Space Station 3D                            & 2002                     & 4169                       \\ \hline
Stalingrad                                  & 2013                     & 10315                      \\ \hline
Step Up 3D                                  & 2010                     & 11051                      \\ \hline
Step Up Revolution                          & 2012                     & 10064                      \\ \hline
Texas Chainsaw 3D                           & 2013                     & 6893                       \\ \hline
The Amazing Spider-Man                      & 2012                     & 8585                       \\ \hline
The Child's Eye                             & 2010                     & 7746                       \\ \hline
The Darkest Hour                            & 2011                     & 8034                       \\ \hline
The Final Destination                       & 2009                     & 8403                       \\ \hline
The Great Gatsby                            & 2013                     & 14295                      \\ \hline
\begin{tabular}[c]{@{}l@{}}The Hobbit:\\ An Unexpected Journey\end{tabular} & 2012                     & 8810                       \\ \hline
\begin{tabular}[c]{@{}l@{}}The Hobbit:\\ The Battle of the Five Armies\end{tabular}   & 2014                     & 14021                      \\ \hline
\begin{tabular}[c]{@{}l@{}}The Hobbit:\\ The Desolation of Smaug\end{tabular}      & 2013                     & 15462                      \\ \hline
The Hole                                    & 2010                     & 8765                       \\ \hline
The Martian                                 & 2015                     & 14075                      \\ \hline
The Three Musketeers                        & 2011                     & 9976                       \\ \hline
The Ultimate Wave Tahiti                    & 2010                     & 4083                       \\ \hline
Ultimate G's                                & 2000                     & 3851                       \\ \hline
Underworld: Awakening                       & 2012                     & 7391                       \\ \hline
X-Men: Days of Future Past                  & 2014                     & 12916          \\ \hline
\hline
\textbf{Overall}                            &                          & \textbf{473042}\\
\hline
\end{tabular}
    \caption{Stereo movies used in our experiments, part 2}
    \label{tab:3dmv-list2}
\end{table}

Following the approach from~\cite{midas}, we use a RefineNet architecture to address the depth estimation problem. For the sake of efficiency, we use Light-Weight RefineNet (LRN)~\cite{lightweightrefinenet}.

The encoders in our experiments are based either on MobileNetv2~\cite{mobilenetv2} or on architectures from the EfficientNet family~\cite{efficientnet}, namely EfficientNet-Lite0 and a set of EfficientNet backbones (B0-B5), pre-trained on the \emph{ImageNet} classification task.

We introduce two modifications to improve efficiency and address stability issues further. Firstly, we replace the layer that maps the encoder output to $256$ channels in the original LRN architecture. Instead, we use 1x1  convolutions that do not change the number of channels. In this block, the output of the encoder layer is fused with the features coming from a deeper layer (see Fig.~\ref{fig:lrn}, top) into the same amount of channels as the encoder yields. Secondly, we noticed that the original Chained Residual Pooling (CRP) blocks cause instabilities in the training process. We fix this issue by replacing summation with averaging by the number of chains (Fig.~\ref{fig:lrn}, bottom). Predictions of our LRN decoder modification are twice as small as target depth maps, so we upscale them to the original resolution via bilinear interpolation.

\section{Stereo movies data}
\label{apdx:stereomovies_data}

Across all the experiments, we use the same collection of datasets for training and testing unless otherwise stated. We train our models on a mixture of RedWeb \cite{redweb}, DIML \cite{diml}, 3D Movies \cite{midas}, and MegaDepth \cite{megadepth} datasets and evaluate them on previously unseen NYUv2 \cite{nyuv2} (654 images), TUM-RGBD \cite{tumrgbd} (1815 images), DIW (74,441 images)\cite{diw}, and KITTI \cite{kitti2012} (161 images) datasets. 
The large-scale DIML dataset covers more than 200 indoor environments and provides absolute depth since it was captured with the Kinect sensor. In contrast with DIML and MegaDepth datasets \cite{megadepth} focus mostly on outdoor static environments, such as architecture and landscapes. MegaDepth was acquired using the SfM technique from crowd-sourced internet images and provides UTS depth. The ReDWeb dataset \cite{redweb} consists of 3600 stereo RGB-D images covering both indoor and outdoor scenarios. It is a small but highly diverse dataset with dynamic scenes, constructed using stereo photos from Flickr. The authors of MIDAS \cite{midas} proposed to use stereo movies for depth estimation models training. 

The original dataset consists of 23 stereo movies and featured video frames from various non-static environments. We use similar data acquisition and processing pipeline, yet we use RAFT \cite{raft} instead of PWCNet \cite{pwcnet} to estimate disparities. Additionally, we extend the list of films with 26 additional stereo movies, totaling 49 movies overall. 
We sample one frame per second from these movies. We leave first and the last 10\% of frames out as they usually belong to opening and closing credits. The disparity is considered valid only in pixels where the discrepancy between left to right and right to left disparities is less than 8 pixels. We leave only those images in the dataset, where the disparity is correct for more than 80\% of pixels and the range of disparity (the difference between maximal and minimal disparities) is more than 8 pixels. Acquired images are highly diverse and contain landscapes, architecture, humans in action, and other scene types. Refer to the tables \ref{tab:3dmv-list1} and \ref{tab:3dmv-list2} for detailed films list.

For our ablation studies, we use the NYUv2 raw \cite{nyuv2} data. We subsample the training set to approximately 150 thousand images and use the original test set of images (654 images).

\section{Metrics}

In this work we use $\delta_{1.25}$ and $\mathrm{rel}$ metrics. These metrics are designed to be used with absolute depth predictions. As our models yield up-to-scale depth predictions, we choose an appropriate scene scale before metrics computation. We select the value that minimizes $L_1$ difference between ground truth and predicted log-depth maps.

\begin{figure*}[h!]
    \centering
    \includegraphics[width=0.85\textwidth]{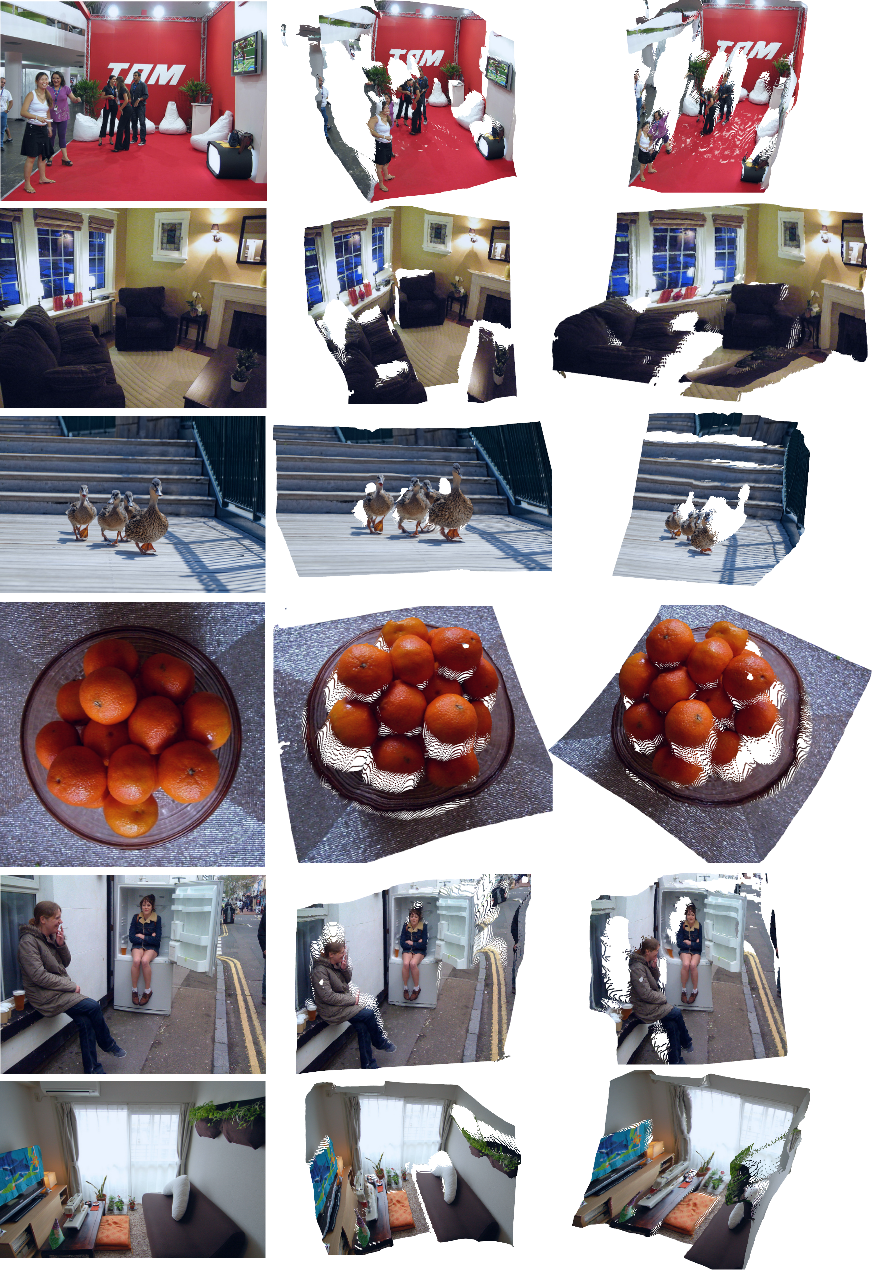}
    \caption{Point cloud 3D reconstructions of images from Microsoft COCO dataset with the use of B5-LRN model. }
    \label{fig:b5-art}
\end{figure*}

\begin{figure*}[h!]
    \centering
    \includegraphics[width=0.7\textwidth]{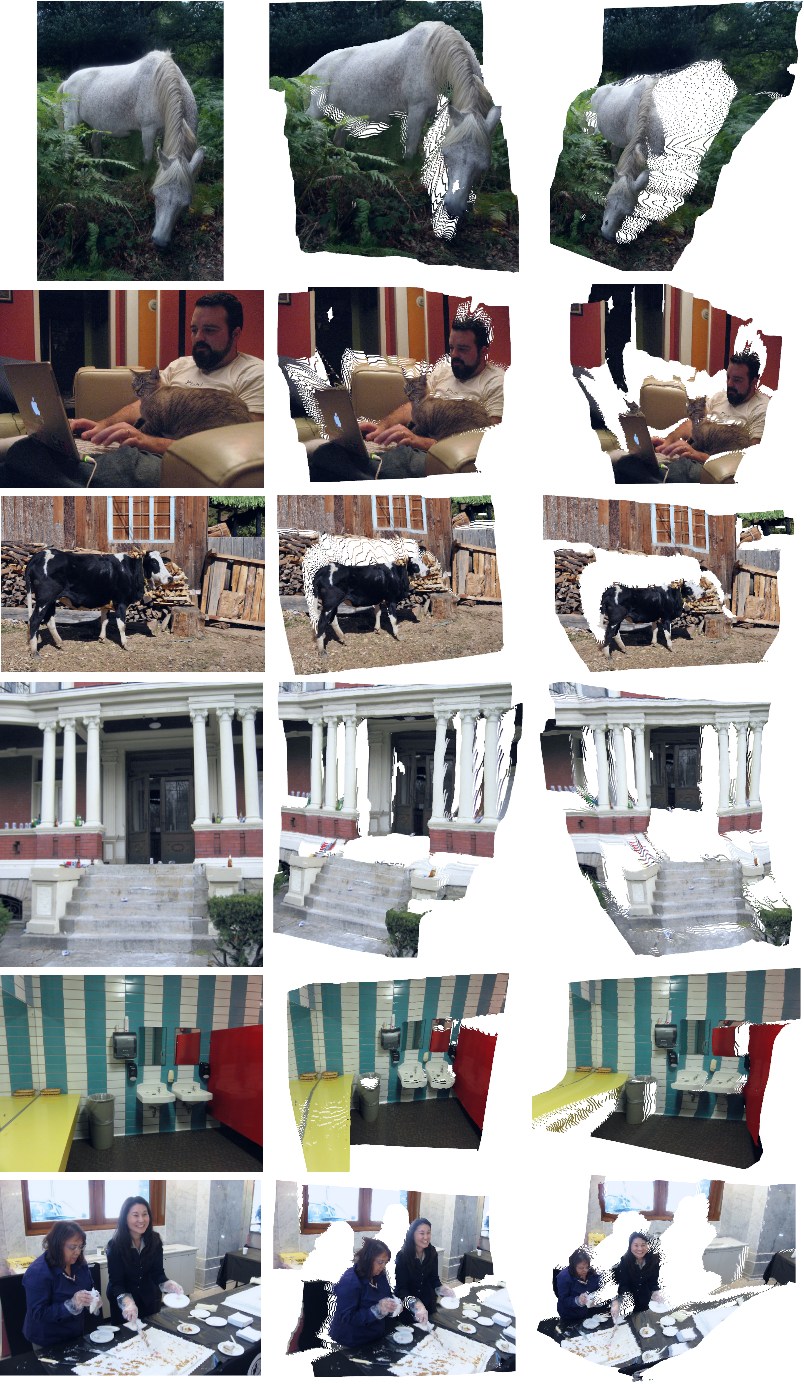}
    \caption{Point cloud 3D reconstructions of images from Microsoft COCO dataset with the use of B5-LRN model. }
    \label{fig:b5-art}
\end{figure*}

\begin{figure*}[h!]
    \centering
    \includegraphics[width=\textwidth]{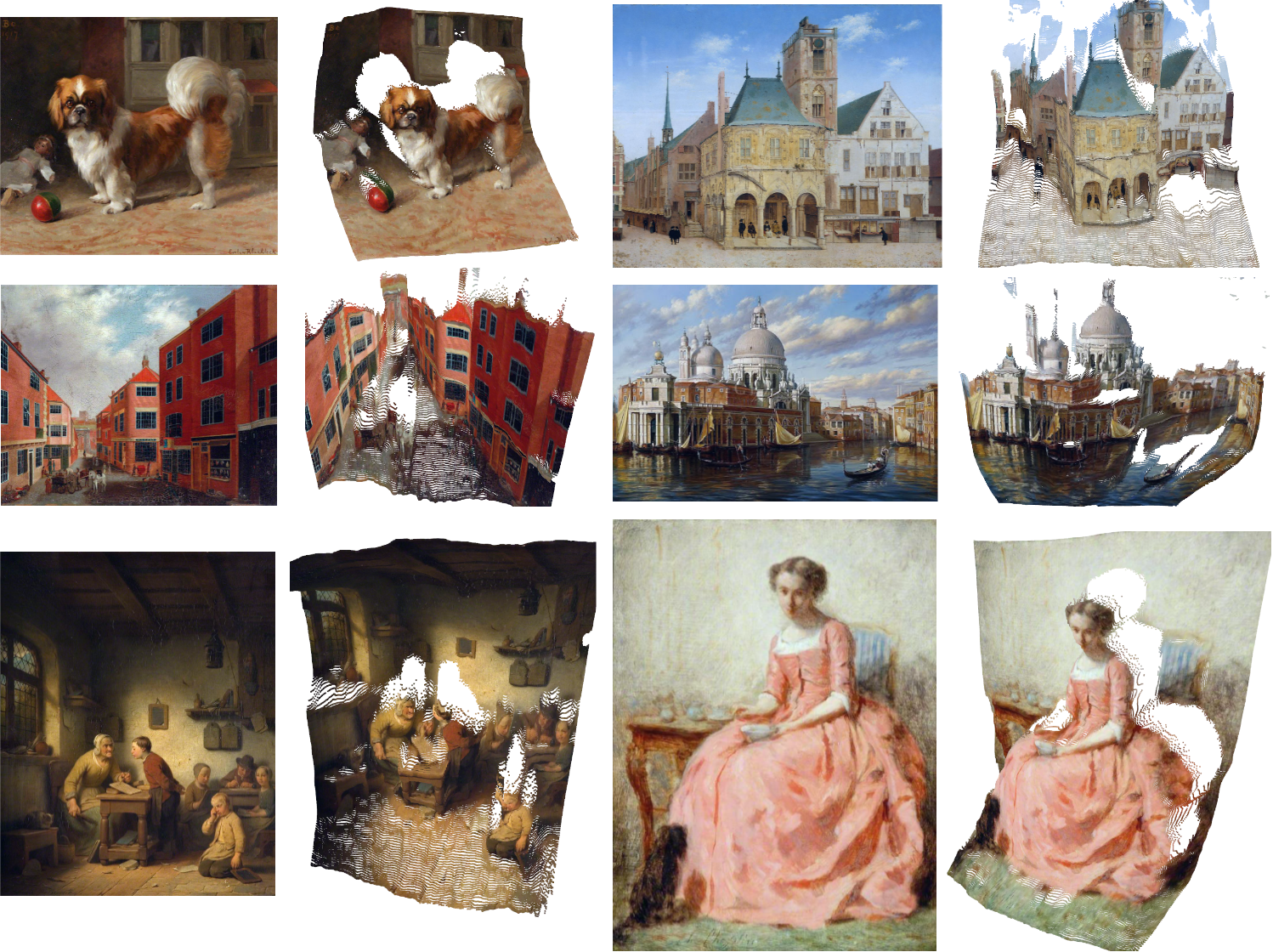}
    \caption{Point cloud 3D reconstructions of paintings with the use of B5-LRN model. }
    \label{fig:b5-art}
\end{figure*}

\begin{figure*}[h!]
    \centering
    \includegraphics[width=\textwidth]{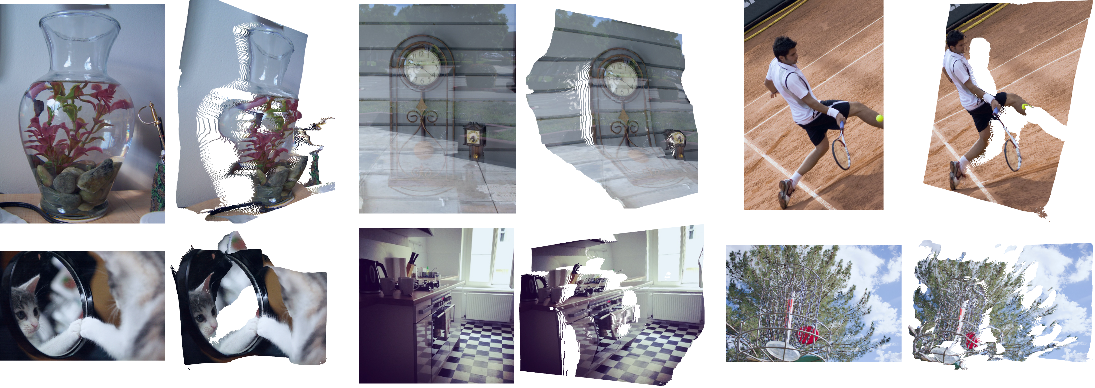}
    \caption{Failure cases for B5-LRN model: reflective and glass surfaces, mirrors, objects with thin edges.}
    \label{fig:b5-art}
\end{figure*}

\end{document}